\documentclass{article}

\usepackage[preprint]{neurips_2026}
\usepackage{enumitem}
\usepackage{tikz}
\usetikzlibrary{arrows.meta, positioning, fit, calc, backgrounds}
\usepackage{bm}
\usepackage{amsmath, amssymb, amsfonts}
\usepackage{bm}
\usepackage{amsthm}
\usepackage{algorithm}
\usepackage{algorithmic}
\usepackage{wrapfig}
\usepackage{booktabs}
\usepackage{multirow}
\usepackage{mdframed}
\usepackage{tcolorbox}
\usepackage{fontawesome5}
\usepackage{bbm}

\usepackage{titlesec}
\tcbset{
  emphblock/.style={
    colback=purple!5!white,
    colframe=purple!75!black,
    fonttitle=\bfseries,
    boxrule=0.8pt,
    arc=4pt,
    left=4pt,
    right=4pt,
    top=4pt,
    bottom=4pt,
  }
}

\titlespacing{\section}{0pt}{6pt}{3pt}

\titlespacing{\subsection}{0pt}{5pt}{2pt}

\titlespacing{\subsubsection}{0pt}{4pt}{2pt}
\titlespacing{\paragraph}{0pt}{4pt}{4pt}

\definecolor{pubblue}{RGB}{220,235,252}
\definecolor{privred}{RGB}{250,224,224}
\definecolor{kvgray}{RGB}{242,242,242}
\definecolor{sanblue}{RGB}{214,231,248}
\definecolor{sharegreen}{RGB}{226,242,226}
\definecolor{attackred}{RGB}{245,214,214}
\definecolor{utilyellow}{RGB}{249,243,210}
\definecolor{mygray}{RGB}{80,80,80}
\newtheorem{remark}{Remark}

\usepackage[utf8]{inputenc} % allow utf-8 input
\usepackage[T1]{fontenc}    % use 8-bit T1 fonts
\usepackage{hyperref}       % hyperlinks
\usepackage{url}            % simple URL typesetting
\usepackage{booktabs}       % professional-quality tables
\usepackage{amsfonts}       % blackboard math symbols
\usepackage{nicefrac}       % compact symbols for 1/2, etc.
\usepackage{microtype}      % microtypography
\usepackage{xcolor}         % colors

\title{LCGuard: Latent Communication Guard for Safe KV Sharing in Multi-Agent Systems}

\author{
\textbf{Sadia Asif}$^{1}$\textsuperscript{\faEnvelope}
\quad
\textbf{Mohammad Mohammadi Amiri}$^{1}$
\quad
\textbf{Momin Abbas}$^{2}$
\\
\textbf{Prasanna Sattigeri}$^{2}$
\quad
\textbf{Karthikeyan Natesan Ramamurthy}$^{2}$
\\[0.6em]
$^{1}$Rensselaer Polytechnic Institute
\\
$^{2}$IBM Research
\\[0.4em]
\faEnvelope\  \texttt{asifs@rpi.edu}
}

\begin{document}

\maketitle

\begin{abstract}
Large language model (LLM)-based multi-agent systems increasingly rely on intermediate communication to coordinate complex tasks. While most existing systems communicate through natural language, recent work shows that latent communication, particularly through transformer key-value (KV) caches, can improve efficiency and preserve richer task-relevant information. However, KV caches also encode contextual inputs, intermediate reasoning states, and agent-specific information, creating an opaque channel through which sensitive content may propagate across agents without explicit textual disclosure. To address this, we introduce \textbf{LCGuard} (Latent Communication Guard), a framework for safe KV-based latent communication in multi-agent LLM systems. LCGuard treats shared KV caches as latent working memory and learns representation-level transformations before cache artifacts are transmitted across agents. We formalize representation-level sensitive information leakage operationally through reconstruction: a shared cache artifact is unsafe if an adversarial decoder can recover agent-specific sensitive inputs from it. This leads to an adversarial training formulation in which the adversary learns to reconstruct sensitive inputs, while LCGuard learns transformations that preserve task-relevant semantics and reduce reconstructable information. Empirical evaluations across multiple model families and multi-agent benchmarks show that LCGuard consistently reduces reconstruction-based leakage and attack success rates while maintaining competitive task performance compared to standard KV-sharing baselines.
\end{abstract}

\section{Introduction}

Large language models (LLMs) are increasingly deployed in \emph{multi-agent systems} (MAS), where multiple specialized agents collaborate through coordination, delegation, and intermediate information exchange to solve complex tasks \cite{li2023camel, wu2024autogen, tran2025multiagent, guo2024llm}. Existing systems predominantly rely on \emph{text-based communication}, in which each agent serializes its internal state into natural language and downstream agents reinterpret that text to continue reasoning \cite{zhu-etal-2025-multiagentbench, qian2025scaling}. While flexible and interpretable, this paradigm is inefficient and lossy: agents repeatedly decode, tokenize, and reconstruct semantic state across communication steps \cite{du2025latent, latentmas}.

Recent work has begun to move beyond text by enabling \emph{latent communication} between agents \cite{shi2026kvcomm, zheng2025thought}. In particular, transformer key-value (KV) caches and intermediate activations have emerged as promising communication substrates \cite{fu2026cache_to_cache, shi2026kvcomm, latentmas}. By directly transferring these representations, agents can avoid redundant computation and preserve richer semantic structure than text-based messages. Such approaches effectively treat KV caches as \emph{shared working memory}, enabling tighter coupling between agents and improving efficiency in multi-stage reasoning.

However, this shift introduces a fundamental and underexplored challenge. Unlike text, which is discrete and externally inspectable, KV caches are high-dimensional, semantically dense representations that encode contextual inputs, intermediate reasoning states, and attention structure. Prior work shows that internal model representations can retain substantial information about their inputs, even when that information is not explicitly decoded \cite{oomerjee2026bottlenecked}. As a result, KV caches may implicitly contain information that never appears in textual outputs. When shared across agents, they form a high-bandwidth representation-space channel whose contents are difficult to inspect or constrain.

This property changes the attack surface of multi-agent systems. When agents exchange KV caches, information is not merely transmitted; it is \emph{embedded}, \emph{transformed}, and \emph{propagated} through internal representations. Sensitive inputs processed by one agent may persist in its KV cache, influence downstream computation, and remain recoverable after multiple stages of interaction. An adversary with access to shared caches, for example through compromised agents, logging infrastructure, or auxiliary models, can exploit this channel by training a decoder to reconstruct underlying inputs. Crucially, this leakage arises entirely at the \emph{representation level} and at \emph{inference time}, without requiring explicit textual disclosure.

Existing approaches are not designed to address this phenomenon. Safety mechanisms in multi-agent systems typically operate over generated outputs or tool actions \cite{guardagent, asif2026information, cheng2026privact, ngong2025protecting}, and therefore do not constrain what is transmitted through latent representations. Meanwhile, prior work on KV-cache security focuses on isolation, eviction, or system-level controls in serving environments \cite{chu2025safekv, luo2026shadow}, rather than on the information content of caches intentionally shared across agents; see Appendix~\ref{app:related_work} for a broader discussion. Consequently, there is currently no principled framework for regulating \emph{what information is retained} in KV representations before they are communicated in multi-agent systems.

This leads to the central challenge:
% \begin{mdframed}[backgroundcolor=blue!5!white,linecolor=blue!75!black,linewidth=1pt]
% \begin{center} 
%     {\sf How can we enable efficient KV-based latent communication in multi-agent systems while limiting the recoverability of agent-specific private information from shared representations?}
%     \end{center} 
% \end{mdframed}

\begin{tcolorbox}[emphblock]
\centering
    How can we enable efficient KV-based latent communication in multi-agent systems while limiting the recoverability of agent-specific private information from shared representations? 
\end{tcolorbox}
% \textcolor{red}{(Momin->Sadia: Added the purple/red box for improved aesthetics. It’s optional—feel free to choose whichever looks best to you (with or without a box))}

% \begin{quote}
% \emph{How can we enable efficient KV-based latent communication in multi-agent systems while limiting the recoverability of agent-specific private information from shared representations?}
% \end{quote}
\begin{wrapfigure}{r}{0.60\linewidth}
    \centering
    \includegraphics[width=\linewidth]{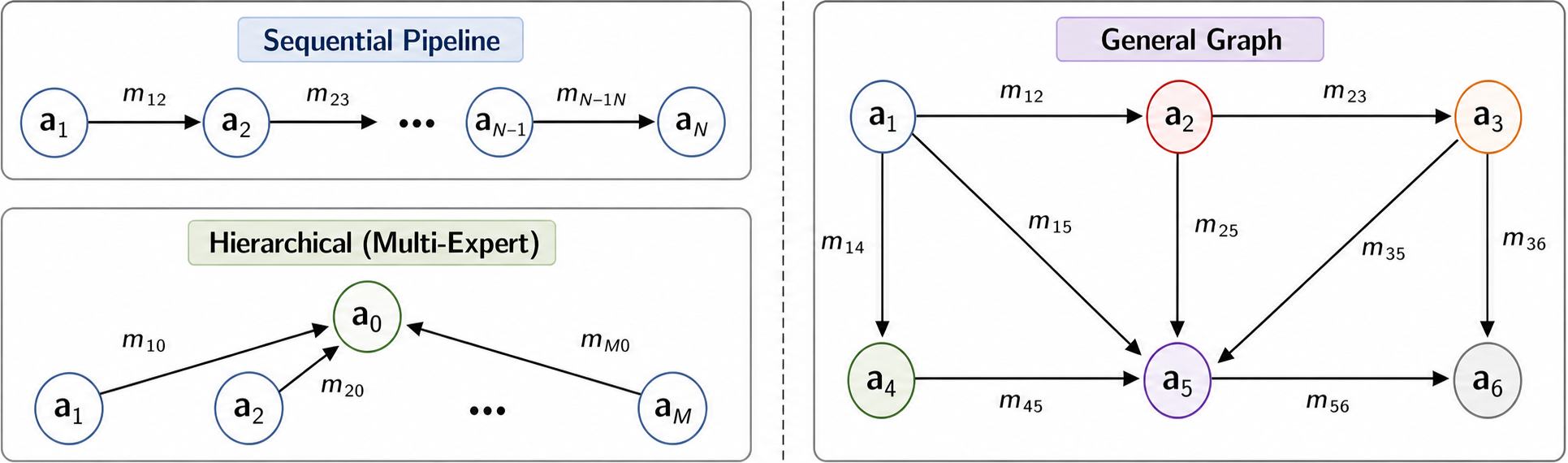}
    \vspace{-10pt}
    \caption{
    Multi-agent communication topologies: sequential, hierarchical, and graph-based. 
    Edges carry KV cache latent artifacts $m_{ij}$.
    }
    \label{fig:topologies}
\end{wrapfigure}

To address this challenge, we introduce \textbf{LCGuard} (Latent Communication Guard), a framework for controlling agent-specific sensitive information flow in KV-based latent communication. LCGuard learns communication functions that transform KV representations before transmission, preserving task-relevant information while limiting the recoverability of sensitive inputs. We formalize representation-level sensitive information leakage in terms of reconstructability: a communicated artifact is considered unsafe if an adversarial decoder can recover agent-specific inputs from it. This yields an adversarial optimization framework in which reconstruction models attempt to infer sensitive inputs, while communication functions are learned to preserve downstream utility and suppress reconstructable information.

We evaluate LCGuard across multiple model families, including Qwen3 (4B, 8B, 14B) \cite{yang2025qwen3}, Gemma-2-9B \cite{gemmateam2024gemma} and LLaMA (3B, 8B) \cite{touvron2023llama2}, under diverse multi-agent settings. Experiments on AgentLeak \cite{elyagoubi2026agentleak}, MAGPIE~\citep{juneja2025magpie} and PrivacyLens \cite{shao2024privacylens} benchmarks show that LCGuard consistently reduces reconstruction-based leakage while maintaining competitive task performance relative to standard KV-sharing baselines.

\paragraph{Contributions.}
Our main contributions are as follows:
\begin{itemize}[leftmargin=*, itemsep=2pt]
    \item We identify \textbf{latent KV communication as an attack surface} in multi-agent LLM systems, where shared working memory can implicitly propagate sensitive information.
    \item We propose \textbf{LCGuard}, a representation-level framework that transforms KV caches before communication to regulate information flow across agents.
    \item We introduce a \textbf{reconstruction-based notion of leakage} that captures the practical recoverability of agent-specific inputs from shared representations.
    \item We formulate safe latent communication as a \textbf{minimax optimization problem} that balances task utility against information exposure.
    \item We empirically demonstrate that LCGuard achieves a favorable \textbf{privacy-utility tradeoff} across multiple models and benchmarks.
\end{itemize}

\section{Problem Setup and Threat Model}
\label{sec:problem}

We consider multi-agent systems that communicate through latent representations derived from Transformer key-value (KV) caches. Such systems may exhibit different interaction patterns, including sequential, hierarchical, or general communication topologies (Figure~\ref{fig:topologies}). Our objective is to formalize this latent communication process at the representation level, with a focus on understanding how transmitted representations simultaneously support task performance and expose agent-specific sensitive information.

Within this setting, we develop a general formulation that abstracts communication as the transmission of learned representations between agents, independent of any specific KV sharing protocol or interaction topology. We first introduce the notation and define the communicated artifacts that mediate interaction. We then formalize reconstruction-based leakage and present a threat model that captures adversarial inference from shared representations. This formulation provides the foundation for the \textbf{LCGuard} framework introduced in Section~\ref{sec:method}.

\subsection{Preliminaries and Notation}
We consider $N$ agents $\{a_1, \dots, a_N\}$, where each agent $a_i$ is modeled as a Transformer parameterized by $\bm{\theta}_i$. For simplicity, we use $a_i$ to denote both the agent and its underlying model throughout the paper. Each agent receives:
(i) a task input $\bm{x}_i \in \mathcal{X}$, where $\mathcal{X}$ denotes the task input space, and  
(ii) an agent-specific (potentially sensitive) input $\bm{s}_i \in \mathcal{S}$, where $\mathcal{S}$ denotes the space of private or agent-specific attributes. These may include user context, retrieved documents, or intermediate outputs that should not be recoverable by other agents or adversaries.

Given these inputs, the agent processes its sequence using a Transformer model and produces internal KV representations. We denote the aggregated key and value tensors for agent $a_i$ as
\begin{equation}
    \bm{K}_i \in \mathbb{R}^{T_i \times d_k}, 
\quad
\bm{V}_i \in \mathbb{R}^{T_i \times d_v},
\end{equation}

where $T_i$ is the number of processed tokens, and $d_k$ and $d_v$ denote the dimensionality of keys and values, respectively.

We assume that communication between agents occurs entirely in the KV latent space, and no explicit text-based information is exchanged. Agents interact by transmitting representations derived from their internal states. For any pair of agents $(a_i,a_j)$, we define the communicated artifact from agent $a_i$ to agent $a_j$ as
\begin{equation}
    \bm{m}_{ij} = g_{ij}(\bm{K}_i, \bm{V}_i),
\end{equation}

where $g_{ij}$ is a learnable transformation function, parameterized by $\bm{\phi}_i$, that determines what information is transmitted from agent $a_i$ to agent $a_j$ (See Appendix \ref{app:comm_architecture} for further details about $g_{ij}$). In particular, if $g_{ij}$ is the identity function, the raw representations $(\bm{K}_i, \bm{V}_i)$ are directly shared. For notational simplicity, we omit the explicit dependence on $\bm{\phi}_i$ and use $g_{ij}$ throughout the paper. We denote the collection of all communicated artifacts in the system as
\begin{equation}
    \bm{\mathcal{M}} = \{ \bm{m}_{ij} \}_{i,j}.
\end{equation}

The downstream task is performed based on these communicated representations, and we define a system-level task loss function
\begin{equation}
    \mathcal{L}_{\mathrm{task}}(\bm{\mathcal{M}})
=
\mathbb{E}
\left[
-\log p(\bm{y} \mid \bm{\mathcal{M}})
\right],
\end{equation}
% \[

% \]
where $\bm{y}$ denotes the task output.

\subsection{Threat Model}

We consider an adversary whose objective is to infer agent-specific sensitive information from communicated latent representations. The adversary operates at the representation level and does not rely on explicit textual outputs, reflecting scenarios where internal communication channels are exposed through shared memory, logging infrastructure, or compromised agents.

\paragraph{Adversarial access.}
The adversary is assumed to observe a subset of communicated artifacts $\bm{\mathcal{M}}_{\mathrm{obs}} \subseteq \bm{\mathcal{M}}$. This captures a range of realistic settings, including partial access to communication links, as well as global access to the entire communication stream.

\paragraph{Adversarial knowledge.}
We assume the adversary has knowledge of the model family. However, the adversary does not have access to the sensitive inputs $\bm{s}_i$ at inference time and must rely solely on the observed artifacts $\bm{\mathcal{M}}_{\mathrm{obs}}$ to perform reconstruction.

\paragraph{Attack Objective.}
The adversary aims to recover $\bm{s}_i$ by minimizing the reconstruction loss 
$\mathcal{L}_{\mathrm{rec}}^{(i)}(\bm{\mathcal{M}}_{\mathrm{obs}})$ using a learned decoder
% The adversary aims to recover $\bm{s}_i$ by minimizing the reconstruction loss 
% $\mathcal{L}_{\mathrm{rec}}^{(i)}(\bm{\mathcal{M}}_{\mathrm{obs}})$
% through a learned decoder $D_i$. For a target agent $a_i$, the adversary employs a decoder
\begin{equation}
    D_i : \bm{\mathcal{M}}_{\mathrm{obs}} \rightarrow \widehat{\bm{s}}_i,
\end{equation}
parameterized by $\bm{\psi}_i$, where $\widehat{\bm{s}}_i$ denotes the reconstructed estimate of the sensitive input $\bm{s}_i$, and $\bm{\mathcal{M}}_{\mathrm{obs}} \subseteq \bm{\mathcal{M}}$ represents the set of observed communicated artifacts.

The adversary aims to infer $\bm{s}_i$ by minimizing the reconstruction loss
\begin{equation}
    \mathcal{L}_{\mathrm{rec}}^{(i)}(\bm{\mathcal{M}}_{\mathrm{obs}})
=
\mathbb{E}
\left[
-\log p(\bm{s}_i \mid \bm{\mathcal{M}}_{\mathrm{obs}})
\right].
\end{equation}
Lower reconstruction loss indicates that the communicated representations received from agent $a_i$ encode more information about $\bm{s}_i$, making it easier for the adversary to recover sensitive inputs. To quantify leakage relative to a baseline, we also define the prior reconstruction loss
\begin{equation}
    \mathcal{L}_{\mathrm{prior}}^{(i)}
=
\mathbb{E}
\left[
-\log p(\bm{s}_i)
\right],
\end{equation}
which corresponds to the best achievable reconstruction performance without access to any communicated artifacts.

\begin{remark}
Leakage can be interpreted through the gap between the reconstruction loss 
$\mathcal{L}_{\mathrm{rec}}^{(i)}(\bm{\mathcal{M}}_{\mathrm{obs}})$ and the prior loss $\mathcal{L}_{\mathrm{prior}}^{(i)}$, under the assumption of a sufficiently strong adversary that approximately minimizes the reconstruction objective. A larger gap indicates that the adversary is more successful in recovering $\bm{s}_i$ from the observed representations, implying higher leakage. Conversely, when the reconstruction loss approaches the prior loss, the adversary gains little advantage from $\bm{\mathcal{M}}_{\mathrm{obs}}$, indicating that the communicated representations provide minimal additional information about $\bm{s}_i$ and thus preserve privacy.
\end{remark}

\paragraph{Local vs. system-level leakage.}
The formulation naturally captures different levels of exposure through the choice of $\bm{\mathcal{M}}_{\mathrm{obs}}$. Local leakage corresponds to the case where the adversary observes a single communicated artifact, i.e., $\bm{\mathcal{M}}_{\mathrm{obs}} = \bm{m}_{ij}$ for some $(i,j)$, measuring leakage at agent $a_i$ from an individual communication link between agents $a_i$  and $a_j$. In contrast, system-level leakage corresponds to $\bm{\mathcal{M}}_{\mathrm{obs}} = \bm{\mathcal{M}}$, where the adversary aggregates information across all communicated artifacts. This distinction highlights how information that may appear weakly encoded in individual transmissions can become significantly more recoverable when combined across multiple communication paths.

This threat model captures inference-time leakage arising from KV based latent communication, where sensitive information may be implicitly encoded in intermediate representations even if it is never explicitly revealed in outputs. By focusing on reconstruction from communicated artifacts, the model reflects a strong and practical adversary that exploits representation-level signals rather than surface-level behavior.

\subsection{Design Objective}

Based on the formulation and threat model above, the goal is to design communication mechanisms that balance task performance captured by $ \mathcal{L}_{\mathrm{task}}(\bm{\mathcal{M}})$  and privacy captured through $\mathcal{L}_{\mathrm{rec}}^{(i)}(\bm{\mathcal{M}}_{\mathrm{obs}})$. Specifically, we seek communication functions $\{g_{ij}\}_{i,j}$ that produce transmitted representations $\bm{\mathcal{M}}$ which are:
(i) informative enough to support downstream task performance i.e., minimizing $\mathcal{L}_{\mathrm{task}}(\bm{\mathcal{M}})$ , and  
(ii) minimally informative about sensitive inputs $\bm{s}_i$ i.e., maximizing $\mathcal{L}_{\mathrm{rec}}^{(i)}(\bm{\mathcal{M}}_{\mathrm{obs}})$ .

This naturally leads to a tradeoff between utility and leakage. Task performance is captured through the task loss $\mathcal{L}_{\mathrm{task}}(\bm{\mathcal{M}})$, while privacy risk is quantified via the reconstruction loss $\mathcal{L}_{\mathrm{rec}}^{(i)}(\bm{\mathcal{M}}_{\mathrm{obs}})$.
The next section introduces \textbf{LCGuard}, which implements this objective as an optimization problem.

\section{Methodology}
\label{sec:method}

We now introduce \textbf{LCGuard}, a framework for controlling agent specific sensitive information flow in latent multi-agent communication. LCGuard learns communication functions $\{g_{ij}\}_{i,j}$ that transform KV representations such that they preserve task-relevant information while limiting the recoverability of sensitive inputs $\{\bm{s}_i\}_i$. To achieve this, LCGuard adopts an adversarial learning framework in which communication functions and reconstruction models are jointly optimized with opposing objectives, enabling explicit control over the utility-privacy tradeoff.

\subsection{Adversarial Learning Formulation}
We model the interaction between communication and reconstruction as a two-player game. For each agent $a_i$, the communication functions $\{g_{ij}\}_{i,j}$ are parameterized by $\bm{\phi}_i$, while an adversarial decoder $D_i$, parameterized by $\bm{\psi}_i$, attempts to reconstruct the sensitive input $\bm{s}_i$ from the observed communicated artifacts $\bm{\mathcal{M}}_{\mathrm{obs}}$. The adversary is trained to minimize the reconstruction loss $\mathcal{L}_{\mathrm{rec}}^{(i)}(\bm{\mathcal{M}}_{\mathrm{obs}})$, thereby extracting as much information as possible from communicated representations. In contrast, the communication functions are trained to make this reconstruction task difficult by maximizing the reconstruction loss $\mathcal{L}_{\mathrm{rec}}^{(i)}(\bm{\mathcal{M}}_{\mathrm{obs}})$  while preserving task-relevant information.

\subsection{Joint Optimization}

The interaction between communication functions and adversarial decoders is formalized as a minimax optimization problem. Specifically, we jointly optimize the communication parameters $\{\bm{\phi}_i\}_i$ and adversarial parameters $\{\bm{\psi}_i\}_i$ as:
\begin{equation}
\min_{\{\bm{\phi}_i\}_i}
\max_{\{\bm{\psi}_i\}_i}
\quad
\beta \,\sum_{i=1}^{N}
\mathcal{L}_{\mathrm{rec}}^{(i)}(\bm{\mathcal{M}}_{\mathrm{obs}})
\;+\;
 \mathcal{L}_{\mathrm{task}}(\bm{\mathcal{M}}),
\label{eq:minimax}
\end{equation}
where $\beta > 0$ controls the tradeoff between task utility and leakage suppression. Larger $\beta$ values place greater emphasis on suppressing leakage, potentially at the cost of task performance, while smaller $\beta$ values prioritize utility. In practice, we treat $\beta$ as a tunable hyperparameter and analyze its effect through privacy-utility tradeoff curves (Appendix~\ref{app:effect-of-beta}).

This objective balances two competing goals: reducing the reconstructability of sensitive inputs while maintaining task performance. The reconstruction term $\mathcal{L}_{\mathrm{rec}}^{(i)}(\bm{\mathcal{M}}_{\mathrm{obs}})$ captures privacy risk under the adversarial model, while the task loss $ \mathcal{L}_{\mathrm{task}}(\bm{\mathcal{M}})$ ensures that communicated representations remain useful for downstream prediction.

\subsection{Training Procedure}

We optimize Eq.~\eqref{eq:minimax} using alternating updates between communication functions and adversarial decoders:

\begin{itemize}[leftmargin=*]
\item \textbf{Adversary step:} Fix $\{\bm{\phi}_i\}_i$ and update $\{\bm{\psi}_i\}_i$ to minimize $\mathcal{L}_{\mathrm{rec}}^{(i)}(\bm{\mathcal{M}}_{\mathrm{obs}})$, strengthening the reconstruction capability of the adversary.
\item \textbf{Communication step:} Fix $\{\bm{\psi}_i\}_i$ and update $\{\bm{\phi}_i\}_i$ to minimize the joint objective, encouraging communication functions to suppress reconstructable information while maintaining task performance.
\end{itemize}

This alternating optimization drives the system toward representations that are maximally useful for the task while being minimally informative about sensitive inputs.

The effectiveness of leakage mitigation depends on the choice of observed artifacts $\bm{\mathcal{M}}_{\mathrm{obs}}$ used in the reconstruction objective, as discussed in Section~\ref{sec:problem}. When $\bm{\mathcal{M}}_{\mathrm{obs}} = \bm{m}_{ij}$ corresponds to a single communication link, the optimization primarily targets leakage localized to that interaction. In contrast, when $\bm{\mathcal{M}}_{\mathrm{obs}} = \bm{\mathcal{M}}$ includes all communicated artifacts, the objective accounts for information aggregated across multiple agents. This enables the learned communication functions to mitigate both local and compositional leakage depending on the adversarial observation setting. Full training details for both local and system-level settings are provided in Appendix~\ref{app:training_algorithm}.

\section{Experiments}
\label{sec:experiments}

We evaluate \textbf{LCGuard} in multi-agent LLM systems that communicate through latent KV representations. Our experiments are designed to answer three questions:
(i) Does KV-based latent communication introduce reconstruction-based privacy leakage?
(ii) Can LCGuard reduce leakage while preserving downstream task performance?
(iii) How does the privacy-utility tradeoff vary across models, benchmarks, and communication topologies?
\subsection{Experimental Setup}
\label{subsec:exp_setup}

We evaluate LCGuard under the reconstruction-based threat model described in Section~\ref{sec:problem}. Our experiments span multiple benchmarks, model families, and communication topologies to assess both task performance and reconstruction-based leakage. We additionally evaluate the inference-time efficiency of LCGuard relative to text-based communication and Vanilla KV sharing in Appendix~\ref{app:efficiency_analysis}.

\paragraph{Benchmarks.}
We consider three multi-agent benchmarks with explicit privacy considerations.
\textbf{PrivacyLens}~\citep{shao2024privacylens} evaluates contextual privacy violations in agent behavior.
\textbf{AgentLeak}~\citep{elyagoubi2026agentleak} focuses on leakage through internal communication channels.
\textbf{MAGPIE}~\citep{juneja2025magpie} studies collaborative settings where private information is directly tied to task success.
\paragraph{Models.}
We evaluate across multiple open-weight LLM backbones, including Qwen3 (4B, 8B, 14B) \cite{yang2025qwen3}, Gemma-2-9B \cite{gemmateam2024gemma} and LLaMA (3B, 8B) \cite{touvron2023llama2}, covering different families and scales.

\paragraph{Baselines.}
We compare LCGuard against baselines spanning different privacy paradigms:
\textbf{Vanilla KV Sharing} (LatentMAS \cite{latentmas} and KVComm \cite{shi2026kvcomm}), which directly transmits latent representations without protection;
\textbf{PrivAct}~\citep{cheng2026privact}, which enforces policy-level privacy constraints;
\textbf{ADAPT}~\citep{fatima2025privacymas}, which injects noise for differential privacy;
and \textbf{Per-Agent LCGuard}, a local variant that applies communication transformations independently at each agent, where $\bm{\mathcal{M}}_{\mathrm{obs}} = \bm{m}_{ij}$.
Our full method is denoted as \textbf{Full-System LCGuard}, where $\bm{\mathcal{M}}_{\mathrm{obs}} = \bm{\mathcal{M}}$.
Detailed descriptions of all baselines, including implementation specifics, are provided in Appendix~\ref{app:baselines}.
\paragraph{Evaluation metrics.}
We evaluate both task performance and sensitive information leakage.
\textbf{Task Accuracy} and \textbf{Helpfulness} measure downstream utility.
\textbf{Privacy Score}, \textbf{Leak Rate}, and \textbf{Attack Success Rate (ASR)} quantify the extent to which sensitive inputs can be recovered from shared representations.
We further analyze \textbf{Reconstruction Difficulty} as a diagnostic metric, which measures the difficulty of reconstructing sensitive inputs from shared representations relative to the prior loss (given in Section~\ref{sec:problem}).
Higher accuracy, helpfulness, and privacy are better, while lower leak rate and ASR indicate stronger protection.
Detailed metric definitions, and ablation studies on adversary strength, the tradeoff parameter $\beta$, and reconstruction difficulty are provided in Appendix~\ref{app:metrics} and Appendix~\ref{app:effect-of-beta}.

 We report results for sequential and hierarchical communication topologies on Qwen3-4B, LLaMA-3.1-8B, and Gemma-2-9B using the PrivacyLens and AgentLeak benchmarks in Section~\ref{sec:results}. Additional results, including other model variants, general graph settings, and MAGPIE evaluations, are deferred to Appendix~\ref{app:additional_results}.

% \paragraph{Evaluation metrics.}
% We evaluate both task performance and sensitive information leakage.
% \textbf{Task Accuracy} and \textbf{Helpfulness} measure downstream utility.
% \textbf{Privacy Score}, \textbf{Leak Rate}, and \textbf{Attack Success Rate (ASR)} quantify the extent to which sensitive inputs can be recovered from communicated representations.
% In addition, we also report \textbf{Reconstruction Difficulty}, which measures the difficulty of inferring sensitive inputs from shared representations relative to the prior, as defined in Section~\ref{sec:problem} and compared with ASR.
% Higher accuracy, helpfulness, reconstruction difficulty, and privacy are better, while lower leak rate and ASR indicate stronger protection.
% Detailed definitions of these metrics are provided in Appendix~\ref{app:metrics}.
 
%  We report results for sequential and hierarchical communication topologies on Qwen3-4B, LLaMA-3.1-8B, and Gemma-2-9B using the PrivacyLens and AgentLeak benchmarks in Section~\ref{sec:results}. Additional results, including other model variants, general graph settings, and MAGPIE evaluations, are provided in Appendix~\ref{app:additional_results}.

\begin{table}[t]
\centering
\caption{
Qwen3-4B results across sequential (Seq) and hierarchical (Hier) communication topologies. Vanilla KV achieves highest helpfulness but suffers from high leakage and ASR. ADAPT improves privacy at the cost of utility. LCGuard provides the best tradeoff, with Full-System LCGuard achieving lowest ASR while maintaining strong helpfulness.}
\small
\setlength{\tabcolsep}{2pt}
\begin{tabular}{llc|l|ccccc}
\toprule
\textbf{Topology} & \textbf{Benchmark} & \textbf{\#Agents} & \textbf{Method} 
& \textbf{Privacy} $\uparrow$ 
& \textbf{Task} $\uparrow$
& \textbf{Help.} $\uparrow$
& \textbf{Leak} $\downarrow$
& \textbf{ASR} $\downarrow$ \\
\midrule

\multirow{10}{*}{Seq}

& \multirow{5}{*}{PrivacyLens}
& \multirow{5}{*}{4}
& Vanilla KV Sharing (LatentMAS)        & 0.420 & \textbf{0.720} & \textbf{0.780} & 0.580 & 0.871 \\
& & 
& PrivAct         & \underline{0.820} & 0.700 & 0.612 & \underline{0.180} & 0.845 \\
& & 
& ADAPT           & \textbf{0.850} & 0.420 & 0.285 & \textbf{0.150} & 0.332 \\
& & 
& Per-Agent LCGuard    & 0.773 & 0.671 & 0.650 & 0.226 & \underline{0.259} \\
& & 
& Full-System LCGuard  & 0.801 & \underline{0.710} & \underline{0.710} & 0.199 & \textbf{0.216} \\

\cmidrule(lr){2-9}

& \multirow{5}{*}{AgentLeak}
& \multirow{5}{*}{3}
& Vanilla KV Sharing (LatentMAS)          & 0.250 & \textbf{0.640} & \textbf{0.760} & 0.750 & 0.880 \\
& & 
& PrivAct         & 0.551 & 0.531 & 0.431 & 0.449 & 0.790 \\
& & 
& ADAPT           & \underline{0.810} & 0.330 & 0.270 & \underline{0.190} & 0.413 \\
& & 
& Per-Agent LCGuard    & 0.762 & 0.582 & 0.630 & 0.238 & \underline{0.291} \\
& & 
& Full-System LCGuard  & \textbf{0.819} & \underline{0.629} & \underline{0.700} & \textbf{0.181} & \textbf{0.259} \\

\midrule

\multirow{10}{*}{Hier}

& \multirow{5}{*}{PrivacyLens}
& \multirow{5}{*}{5}
&Vanilla KV Sharing (LatentMAS)          & 0.395 & \textbf{0.710} & \textbf{0.755} & 0.610 & 0.873 \\
& & 
& PrivAct         & 0.798 & 0.690 & 0.485 & 0.202 & 0.820 \\
& & 
& ADAPT           & \underline{0.833} & 0.408 & 0.270 & \underline{0.167} & 0.301 \\
& & 
& Per-Agent LCGuard    & 0.781 & 0.680 & 0.600 & 0.219 & \underline{0.272} \\
& & 
& Full-System LCGuard  & \textbf{0.835} & \underline{0.700} & \underline{0.685} & \textbf{0.165} & \textbf{0.238} \\

\cmidrule(lr){2-9}

& \multirow{5}{*}{AgentLeak}
& \multirow{5}{*}{3}
& Vanilla KV Sharing (LatentMAS)          & 0.230 & \textbf{0.630} & \textbf{0.735} & 0.770 & 0.891 \\
& & 
& PrivAct         & 0.534 & 0.562 & 0.371 & 0.466 & 0.752 \\
& & 
& ADAPT           & \underline{0.800} & 0.312 & 0.255 & \underline{0.200} & 0.394 \\
& & 
& Per-Agent LCGuard    & 0.762 & 0.572 & 0.610 & 0.238 & \underline{0.265} \\
& & 
& Full-System LCGuard  & \textbf{0.823} & \underline{0.591} & \underline{0.675} & \textbf{0.177} & \textbf{0.224} \\

\bottomrule
\end{tabular}
\label{tab:qwen4b_main}
\end{table}

\begin{table}[t]
\centering
\caption{
Gemma-9B results across sequential (Seq) and hierarchical (Hier) communication topologies. While stronger baselines increase both utility and leakage, LCGuard consistently reduces ASR and leakage while preserving high helpfulness. Full-System LCGuard performs best overall for privacy-utility tradeoff.
}
\small
\setlength{\tabcolsep}{2pt}
\begin{tabular}{llc|l|ccccc}
\toprule
\textbf{Topology} & \textbf{Benchmark} & \textbf{\#Agents} & \textbf{Method} 
& \textbf{Privacy} $\uparrow$ 
& \textbf{Task} $\uparrow$
& \textbf{Help.} $\uparrow$
& \textbf{Leak} $\downarrow$
& \textbf{ASR} $\downarrow$ \\
\midrule

\multirow{10}{*}{Seq}

& \multirow{5}{*}{PrivacyLens}
& \multirow{5}{*}{4}
& Vanilla KV Sharing (LatentMAS)  & 0.440 & \textbf{0.740} & \textbf{0.805} & 0.560 & 0.885 \\
& & 
& PrivAct                         & \underline{{0.835}} & 0.715 & 0.640 & \underline{0.165} & 0.830 \\
& & 
& ADAPT                           & \textbf{0.870} & 0.440 & 0.305 & \textbf{0.130} & 0.315 \\
& & 
& Per-Agent LCGuard               & 0.790 & 0.690 & 0.675 & 0.210 & \underline{0.245} \\
& & 
& Full-System LCGuard             & 0.825 & \underline{0.725} & \underline{0.735} & 0.175 & \textbf{0.205} \\

\cmidrule(lr){2-9}

& \multirow{5}{*}{AgentLeak}
& \multirow{5}{*}{3}
& Vanilla KV Sharing (LatentMAS)  & 0.270 & \textbf{0.660} & \textbf{0.785} & 0.730 & 0.895 \\
& & 
& PrivAct                         & 0.582 & 0.560 & 0.455 & 0.418 & 0.770 \\
& & 
& ADAPT                           & \underline{0.830} & 0.350 & 0.285 & \underline{0.170} & 0.400 \\
& & 
& Per-Agent LCGuard               & 0.785 & 0.610 & 0.660 & 0.215 & \underline{0.275} \\
& & 
& Full-System LCGuard             & \textbf{0.845} & \underline{0.650} & \underline{0.720} & \textbf{0.155} & \textbf{0.235} \\

\midrule

\multirow{10}{*}{Hier}

& \multirow{5}{*}{PrivacyLens}
& \multirow{5}{*}{5}
& Vanilla KV Sharing (LatentMAS)  & 0.413 & \textbf{0.730} & \textbf{0.790} & 0.587 & 0.890 \\
& & 
& PrivAct                         & 0.821 & 0.705 & 0.520 & 0.179 & 0.805 \\
& & 
& ADAPT                           & \underline{0.855} & 0.430 & 0.290 & \underline{0.145} & 0.295 \\
& & 
& Per-Agent LCGuard               & 0.800 & 0.695 & 0.645 & 0.200 & \underline{0.260} \\
& & 
& Full-System LCGuard             & \textbf{0.860} & \underline{0.715} & \underline{0.720} & \textbf{0.140} & \textbf{0.225} \\

\cmidrule(lr){2-9}

& \multirow{5}{*}{AgentLeak}
& \multirow{5}{*}{5}
& Vanilla KV Sharing (LatentMAS)  & 0.250 & \textbf{0.650} & \textbf{0.765} & 0.750 & 0.900 \\
& & 
& PrivAct                         & 0.560 & 0.580 & 0.390 & 0.440 & 0.740 \\
& & 
& ADAPT                           & \underline{0.822} & 0.330 & 0.270 & \underline{0.178} & 0.380 \\
& & 
& Per-Agent LCGuard               & 0.786 & 0.600 & 0.640 & 0.215 & \underline{0.254} \\
& & 
& Full-System LCGuard             & \textbf{0.850} & \underline{0.630} & \underline{0.705} & \textbf{0.150} & \textbf{0.215} \\

\bottomrule
\end{tabular}
\label{tab:gemma9b_main}
\end{table}

\begin{table}[t]
\centering
\caption{
LLaMA-8B results across sequential (Seq) and hierarchical (Hier) communication topologies. Vanilla KV maximizes helpfulness but remains vulnerable. ADAPT reduces leakage with large utility loss. LCGuard achieves a balanced tradeoff, with Full-System LCGuard yielding lowest ASR.
}
\small
\setlength{\tabcolsep}{2pt}
\begin{tabular}{llc|l|ccccc}
\toprule
\textbf{Topology} & \textbf{Benchmark} & \textbf{\#Agents} & \textbf{Method} 
& \textbf{Privacy} $\uparrow$ 
& \textbf{Task} $\uparrow$
& \textbf{Help.} $\uparrow$
& \textbf{Leak} $\downarrow$
& \textbf{ASR} $\downarrow$ \\
\midrule

\multirow{10}{*}{Seq}

& \multirow{5}{*}{PrivacyLens}
& \multirow{5}{*}{4}
& Vanilla KV Sharing (LatentMAS)  & 0.405 & \textbf{0.755} & \textbf{0.835} & 0.595 & 0.760 \\
& & 
& PrivAct                         & 0.835 & 0.725 & 0.455 & 0.165 & 0.735 \\
& & 
& ADAPT                           & \textbf{0.875} & 0.455 & 0.330 & \textbf{0.135} & 0.395 \\
& & 
& Per-Agent LCGuard               & 0.805 & 0.695 & 0.635 & 0.195 & \underline{0.330} \\
& & 
& Full-System LCGuard             & \underline{0.845} & \underline{0.735} & \underline{0.691} & \underline{0.155} & \textbf{0.285} \\

\cmidrule(lr){2-9}

& \multirow{5}{*}{AgentLeak}
& \multirow{5}{*}{3}
& Vanilla KV Sharing (LatentMAS)  & 0.235 & \textbf{0.675} & \textbf{0.835} & 0.765 & 0.760 \\
& & 
& PrivAct                         & 0.565 & 0.555 & 0.455 & 0.435 & 0.735 \\
& & 
& ADAPT                           & \underline{0.840} & 0.355 & 0.330 & \underline{0.160} & 0.395 \\
& & 
& Per-Agent LCGuard               & 0.800 & 0.605 & 0.635 & 0.200 & \underline{0.330} \\
& & 
& Full-System LCGuard             & \textbf{0.860} & \underline{0.655} & \underline{0.691} & \textbf{0.140} & \textbf{0.285} \\

\midrule

\multirow{10}{*}{Hier}

& \multirow{5}{*}{PrivacyLens}
& \multirow{5}{*}{5}
& Vanilla KV Sharing (LatentMAS)  & 0.385 & \textbf{0.745} & \textbf{0.815} & 0.615 & 0.785 \\
& & 
& PrivAct                         & 0.815 & 0.710 & 0.440 & 0.185 & 0.750 \\
& & 
& ADAPT                           & \underline{0.860} & 0.440 & 0.315 & \underline{0.140} & 0.420 \\
& & 
& Per-Agent LCGuard               & 0.810 & 0.690 & 0.641 & 0.190 & \underline{0.355} \\
& & 
& Full-System LCGuard             & \textbf{0.875} & \underline{0.720} & \underline{0.703} & \textbf{0.125} & \textbf{0.305} \\

\cmidrule(lr){2-9}

& \multirow{5}{*}{AgentLeak}
& \multirow{5}{*}{5}
& Vanilla KV Sharing (LatentMAS)  & 0.215 & \textbf{0.665} & \textbf{0.815} & 0.785 & 0.785 \\
& & 
& PrivAct                         & 0.545 & 0.575 & 0.440 & 0.455 & 0.750 \\
& & 
& ADAPT                           & \underline{0.835} & 0.335 & 0.315 & \underline{0.165} & 0.420 \\
& & 
& Per-Agent LCGuard               & 0.805 & 0.595 & 0.641 & 0.195 & \underline{0.355} \\
& & 
& Full-System LCGuard             & \textbf{0.870} & \underline{0.625} & \underline{0.703} & \textbf{0.130} & \textbf{0.305} \\

\bottomrule
\end{tabular}
\label{tab:llama8b_main}
\end{table}

\begin{figure*}[t]
    \centering
    \includegraphics[width=\textwidth]{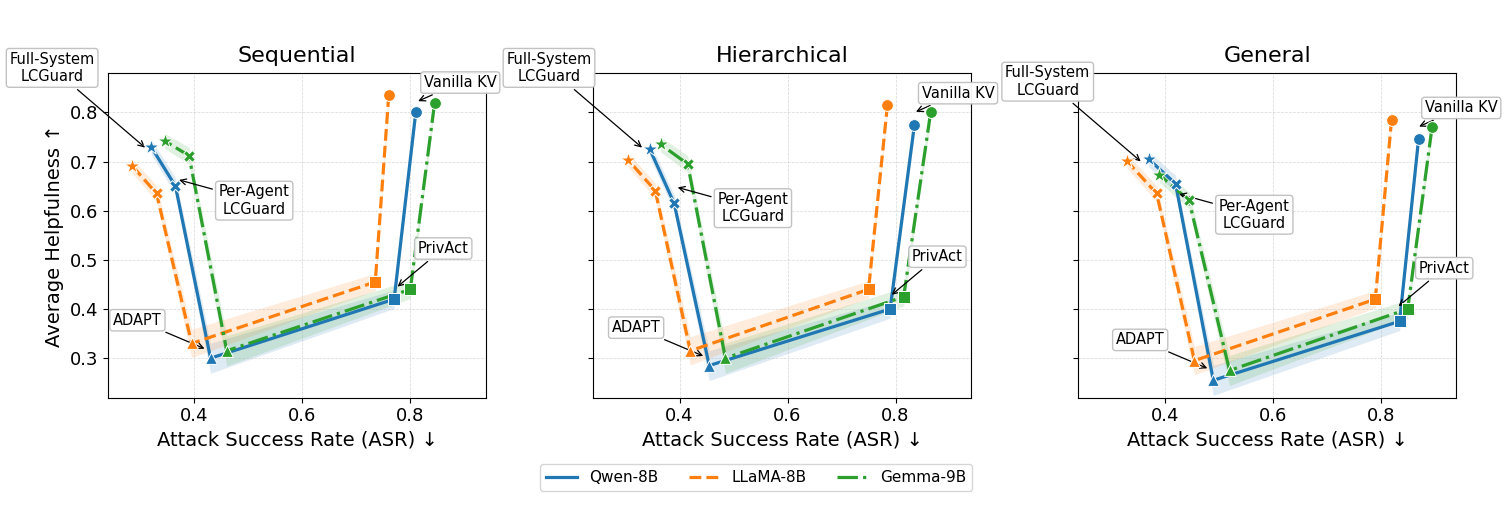}
    \caption{
ASR-helpfulness tradeoff across sequential ($4$ agents), hierarchical ($5$ agents), and general ($5$ agents graph detailed in Appendix~\ref{par:agent_rol}) communication topologies for Qwen-8B, LLaMA-8B, and Gemma-9B. The desirable region is the upper-left, corresponding to low ASR and high helpfulness. Vanilla KV sharing preserves helpfulness but remains highly vulnerable, ADAPT lowers ASR by heavily degrading helpfulness, and LCGuard shifts the frontier toward lower ASR while maintaining substantially higher helpfulness.
}
    \label{fig:asr_helpfulness_tradeoff}
\end{figure*}

\section{Results}
\label{sec:results}

Tables~\ref{tab:qwen4b_main}, \ref{tab:gemma9b_main}, and \ref{tab:llama8b_main} report results across PrivacyLens and AgentLeak under sequential and hierarchical communication. Figure~\ref{fig:asr_helpfulness_tradeoff} summarizes the ASR-helpfulness tradeoff across communication topologies and model families. Across all settings, a consistent pattern emerges: raw KV sharing maximizes utility but exposes substantial private information, while LCGuard significantly reduces reconstruction risk with only moderate utility degradation.

\subsection{Overall Privacy-Utility Tradeoff}

Vanilla KV Sharing achieves the highest task accuracy and helpfulness across all models and benchmarks. For example, in the Qwen3-4B sequential PrivacyLens setting (Table~\ref{tab:qwen4b_main}), it achieves helpfulness $0.780$ and task accuracy $0.720$, while in Gemma-9B (Table~\ref{tab:gemma9b_main}) helpfulness increases further to $0.805$. However, this utility comes with consistently high leakage: ASR remains extremely high, reaching $0.871$ (Qwen3-4B, Table~\ref{tab:qwen4b_main}), $0.885$ (Gemma-9B, Table~\ref{tab:gemma9b_main}), and up to $0.900$ in hierarchical AgentLeak.

LCGuard substantially shifts this tradeoff. Full-System LCGuard consistently achieves the lowest ASR across all configurations. For instance, in Qwen3-4B sequential PrivacyLens (Table~\ref{tab:qwen4b_main}), ASR drops from $0.871$ (Vanilla KV) to $0.216$, while maintaining helpfulness at $0.710$. Similarly, in Gemma-9B (Table~\ref{tab:gemma9b_main}), ASR reduces from $0.885$ to $0.205$ with helpfulness remaining high at $0.735$. Across all tables, this corresponds to an approximate $65$-$75\%$ reduction in ASR while preserving most of the task performance.

The baselines highlight complementary failure modes. ADAPT achieves strong privacy but collapses utility: in Qwen3-4B sequential PrivacyLens (Table~\ref{tab:qwen4b_main}), helpfulness drops to $0.285$ despite low ASR ($0.332$). Similar degradation is observed across Gemma-9B and LLaMA-8B (Tables~\ref{tab:gemma9b_main} and \ref{tab:llama8b_main}), confirming that noise-based protection removes both sensitive and task-relevant information. In contrast, PrivAct improves privacy scores (e.g., $0.820$ in Table~\ref{tab:qwen4b_main}) but leaves ASR largely unchanged ($0.845$), indicating that output-level constraints do not eliminate latent leakage.

\subsection{System-Level vs. Local Protection}

Comparing Per-Agent LCGuard with Full-System LCGuard reveals the importance of system-level optimization. While Per-Agent LCGuard already reduces ASR significantly, Full-System LCGuard consistently achieves further improvements. For example, in Qwen3-4B hierarchical AgentLeak (Table~\ref{tab:qwen4b_main}), ASR decreases from $0.265$ (Per-Agent) to $0.224$ (Full-System), while privacy improves from $0.762$ to $0.823$. Similar gains are observed for Gemma-9B (Table~\ref{tab:gemma9b_main}), where ASR reduces from $0.254$ to $0.215$. These improvements indicate that leakage is not purely local to individual communication edges. Instead, sensitive information can re-emerge after aggregation or transformation across agents. Optimizing sanitization jointly across the system better captures this compositional leakage, leading to stronger protection.

\subsection{Effect of Communication Topology}

Topology influences both privacy and utility. Sequential communication exhibits a more controlled flow of information, while hierarchical communication introduces aggregation points that increase exposure. This is visible in Vanilla KV results: for Qwen3-4B AgentLeak, ASR increases from $0.880$ (sequential) to $0.891$ (hierarchical) (Table~\ref{tab:qwen4b_main}).

Despite this increased difficulty, the relative ordering of methods remains stable across all tables. Vanilla KV consistently occupies the high-helpfulness but high-ASR region, ADAPT remains low-ASR but low-utility, and LCGuard achieves a favorable middle ground. Figure~\ref{fig:asr_helpfulness_tradeoff} further shows that LCGuard moves toward the upper-left region (low ASR, high helpfulness) across sequential, hierarchical, and general graph topologies, indicating robustness to communication structure.

\subsection{Effect of Model Family}

Model capacity influences both utility and leakage. Larger models such as Gemma-9B and LLaMA-8B achieve higher helpfulness under Vanilla KV (e.g., $0.805$ in Table~\ref{tab:gemma9b_main} and $0.835$ in Table~\ref{tab:llama8b_main}), but also exhibit high ASR (up to $0.900$). This suggests that stronger latent representations encode more recoverable information. LCGuard remains effective across all model families. In LLaMA-8B sequential PrivacyLens (Table~\ref{tab:llama8b_main}), Full-System LCGuard reduces ASR from $0.760$ to $0.285$ while maintaining helpfulness at $0.691$. Similar trends hold across all settings, indicating that LCGuard generalizes across architectures and scales.

\subsection{Key Insights}

The results support four key insights. First, KV-based communication provides strong task benefits but creates a high-capacity reconstruction channel, as evidenced by consistently high ASR values in Tables~\ref{tab:qwen4b_main}-\ref{tab:llama8b_main}. Second, output-level privacy methods such as PrivAct are insufficient for controlling latent leakage, since ASR remains high despite improved privacy scores. Third, noise-based methods such as ADAPT reduce leakage but significantly degrade utility. Finally, LCGuard achieves the best privacy-utility tradeoff by directly transforming shared KV representations, with Full-System LCGuard consistently achieving the lowest ASR across all reported settings.

\subsection{Reconstruction Difficulty Across Agents}
\begin{wrapfigure}{r}{0.7\textwidth}
\vspace{-10pt}
\centering
\includegraphics[width=0.7\textwidth]{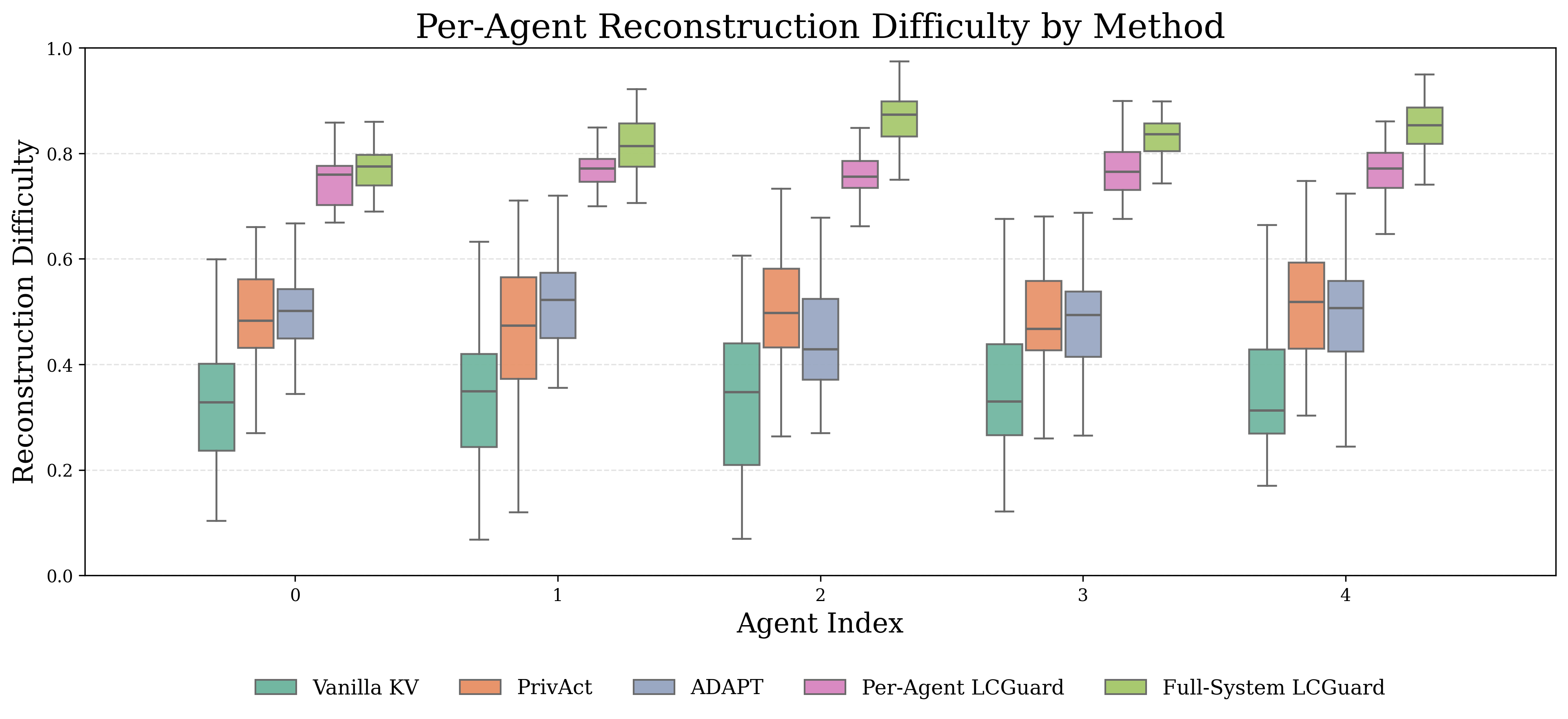}
\vspace{-10pt}
\caption{
Per-agent reconstruction difficulty across methods in a sequential 5-agent setting for Qwen3-4B model. LCGuard consistently increases reconstruction difficulty at all positions in the pipeline, with the largest gains observed at later agents where compositional leakage accumulates.
}
\label{fig:per_agent_rd}
\vspace{-10pt}
\end{wrapfigure}
To better understand how sensitive information propagates across agents, we analyze \emph{per-agent reconstruction difficulty}, given in Eq. \ref{equ:rec_difficulty} along the communication pipeline (Detailed analysis given in Appendix \ref{app:reconstruction_difficulty_ablation}. Figure~\ref{fig:per_agent_rd} shows the distribution of reconstruction difficulty across agents for different methods.

A clear trend emerges: reconstruction difficulty increases along the agent chain for LCGuard, particularly for the full-system variant. In contrast, Vanilla KV exhibits consistently low reconstruction difficulty across agents, indicating that sensitive information remains easily recoverable throughout the pipeline.\ The gap between Per-Agent and Full-System LCGuard becomes more pronounced at deeper agents, highlighting that leakage is not purely local but accumulates through multi-hop transformations. This supports our design choice of system-level optimization.
\section{Conclusion and Future Work}
We introduced \textbf{LCGuard}, a representation-level framework for regulating sensitive information flow in KV-based latent communication for multi-agent LLM systems. LCGuard learns communication functions that transform shared KV artifacts before transmission, preserving task-relevant information while reducing the recoverability of agent-specific sensitive inputs. By framing leakage through reconstruction from communicated representations, our formulation captures inference-time privacy risks that are not addressed by output-level safeguards. Across multiple model families, benchmarks, and communication topologies, LCGuard consistently reduces leakage and ASR while maintaining competitive task accuracy and helpfulness. Several directions remain open. First, extending LCGuard to \emph{heterogeneous multi-agent systems}, where agents may use different architectures or incompatible latent spaces, would make the framework more applicable to realistic deployments. Second, adapting LCGuard to \emph{multimodal agents} that exchange text, vision, audio, or tool-derived representations would broaden its scope beyond language-only settings. Finally, learning \emph{adaptive communication policies} that decide when, what, and how much latent information should be shared could further improve efficiency, robustness, and privacy in large-scale multi-agent systems.

\clearpage
\bibliographystyle{plain}
\bibliography{refs}

{
% \small

%%%%%%%%%%%%%%%%%%%%%%%%%%%%%%%%%%%%%%%%%%%%%%%%%%%%%%%%%%%%

\appendix
% \appendix

\section{Appendix}
\subsection{Training Algorithm}
\label{app:training_algorithm}

We train LCGuard using a minimax optimization procedure that alternates between an adversarial decoder (attacker) and a communication transformation function (defender). The objective follows Eq.~\eqref{eq:minimax}, where the decoder seeks to reconstruct sensitive inputs $\bm{s}_i$ from observed representations $\bm{\mathcal{M}}_{\mathrm{obs}}$, while the communication function learns to preserve task utility while suppressing reconstructable information.
Algorithms~\ref{alg:full_system_lcguard} and~\ref{alg:per_agent_lcguard} summarize the training procedures for the full-system and per-agent variants of LCGuard, respectively.

\begin{algorithm}[t]
\caption{Full-System LCGuard Training}
\label{alg:full_system_lcguard}
\begin{algorithmic}[1]
\STATE \textbf{Inputs:} dataset $\mathcal{D}$, communication functions $\{g_{ij}\}_{i,j}$, decoders $\{D_{i}\}_i$, tradeoff $\beta$, attacker steps $k$
\STATE \textbf{Initialize} $\{\phi_{ij}\ \text{(for } g_{ij}\text{)},\ \psi_i\ \text{(for } D_i\text{)}\}_{i}$

\FOR{each training iteration}
    \STATE Sample minibatch 
    $\mathcal{B} = \{(\{\bm{x}_i^{(b)}, \bm{s}_i^{(b)}\}_{i=1}^N, \bm{y}^{(b)})\}_{b=1}^B$
    
    \STATE \textbf{Forward pass (agents):}
    \STATE Compute raw KV representations 
    $\{(\bm{K}_i^{(b)}, \bm{V}_i^{(b)})\}_{i=1}^N$ for each $b$
    
    \STATE \textbf{Communication transformation:}
      \[
    \bm{\mathcal{M}}^{(b)} \leftarrow g_{ij}\big(\{(\bm{K}_i^{(b)}, \bm{V}_i^{(b)})\}_{i=1}^N\big)
    \]
    \STATE Set $\bm{\mathcal{M}}_{\mathrm{obs}}^{(b)} = \bm{\mathcal{M}}^{(b)}$
    
    \FOR{$t = 1$ to $k$}
        \STATE \textbf{Decoder update (attacker):}
        \[
        \psi_i \leftarrow \psi_i
        -
        \eta_{\psi}
        \nabla_{\psi_i}
        \sum_{b=1}^{B}
        \mathcal{L}_{\mathrm{rec}}^{(i)}
        \!\left(
        D_{i,\psi_i}(\bm{\mathcal{M}}_{\mathrm{obs}}^{(b)}),
        \bm{s}_i^{(b)}
        \right)
        \]
    \ENDFOR
    
    \STATE \textbf{Communication update (defender):}
    \[
    \phi_{i}\
    \leftarrow
    \phi_{i}
    -
    \eta_{\phi}
    \nabla_{\phi_{i}}
    \sum_{b=1}^{B}
    \Bigg[
    \mathcal{L}_{\mathrm{task}}
    \!\left(\bm{\mathcal{M}}^{(b)}\right)
    -
    \beta
    \sum_{i=1}^{N}
    \mathcal{L}_{\mathrm{rec}}^{(i)}
    \!\left(
    \bm{\mathcal{M}}_{\mathrm{obs}}^{(b)},
    \bm{s}_i^{(b)}
    \right)
    \Bigg]
    \]
\ENDFOR
\end{algorithmic}
\end{algorithm}
% \begin{algorithm}[t]
% \caption{Full-System LCGuard Training}
% \label{alg:full_system_lcguard}
% \begin{algorithmic}[1]
% \STATE \textbf{Inputs:} dataset $\mathcal{D}$, communication functions $g_{\phi}$, decoder $D_{\psi}$, tradeoff $\beta$, steps $k$
% \STATE \textbf{Initialize} $\phi, \psi$

% \FOR{each training iteration}
%     \STATE Sample minibatch $\mathcal{B} = \{(\{\bm{x}_i^{(b)}, \bm{s}_i^{(b)}\}_{i=1}^N, \bm{y}^{(b)})\}_{b=1}^B$
    
%     \STATE \textbf{Forward pass (agents):}
%     \STATE Compute raw KV representations $\{(\bm{K}_i^{(b)}, \bm{V}_i^{(b)})\}_{i=1}^N$
    
%     \STATE \textbf{Communication transformation:}
%     \[
%     \bm{\mathcal{M}}^{(b)} \leftarrow g_{\phi}\big(\{(\bm{K}_i^{(b)}, \bm{V}_i^{(b)})\}_{i=1}^N\big)
%     \]
%     \STATE Set $\bm{\mathcal{M}}_{\mathrm{obs}}^{(b)} = \bm{\mathcal{M}}^{(b)}$
    
%     \FOR{$t = 1$ to $k$}
%         \STATE \textbf{Decoder update (attacker):}
%         \[
%         \psi \leftarrow \psi - \eta_{\psi} \nabla_{\psi} \sum_{i=1}^{N} 
%         \mathcal{L}_{\mathrm{rec}}^{(i)}(\bm{\mathcal{M}}_{\mathrm{obs}}, \bm{s}_i)
%         \]
%     \ENDFOR
    
%     \STATE \textbf{Communication update (defender):}
%     \[
%     \phi \leftarrow \phi - \eta_{\phi} \nabla_{\phi} 
%     \Big(
%     \mathcal{L}_{\mathrm{task}}(\bm{\mathcal{M}})
%     -
%     \beta \sum_{i=1}^{N} \mathcal{L}_{\mathrm{rec}}^{(i)}(\bm{\mathcal{M}}_{\mathrm{obs}}, \bm{s}_i)
%     \Big)\]
% \ENDFOR
% \end{algorithmic}
% \end{algorithm}

\begin{algorithm}[t]
\caption{Per-Agent LCGuard Training}
\label{alg:per_agent_lcguard}
\begin{algorithmic}[1]
\STATE \textbf{Inputs:} dataset $\mathcal{D}$, communication functions $\{g_{ij}\}_{i,j}$, decoders $\{D_{i}\}_i$, tradeoff $\beta$, attacker steps $k$
\STATE \textbf{Initialize} $\{\phi_{ij}\ \text{(for } g_{ij}\text{)},\ \psi_i\ \text{(for } D_i\text{)}\}_{i}$

\FOR{each training iteration}
    \FOR{$i = 1$ to $N$}
        \STATE Sample minibatch $\mathcal{B}_i = \{(\bm{x}_i^{(b)}, \bm{s}_i^{(b)}, \bm{y}^{(b)})\}_{b=1}^B$
        
        \STATE \textbf{Forward pass (agent $a_i$):}
        \[
        (\bm{K}_i^{(b)}, \bm{V}_i^{(b)}) \leftarrow \text{Agent}_i(\bm{x}_i^{(b)})
        \]
        
        \STATE \textbf{Local communication transformation:}
        \[
        \bm{m}_{ij}^{(b)} \leftarrow g_{ij}(\bm{K}_i^{(b)}, \bm{V}_i^{(b)})
        \]
        
        \STATE Construct $\bm{\mathcal{M}}_{\mathrm{obs}}^{(b)}$ from $\bm{m}_{ij}^{(b)}$
        
        \FOR{$t = 1$ to $k$}
            \STATE \textbf{Decoder update (attacker for $a_i$):}
            \[
            \psi_i \leftarrow \psi_i - \eta_{\psi} \nabla_{\psi_i} 
            \mathcal{L}_{\mathrm{rec}}^{(i)}
        \!\left(
        D_{i,\psi_i}(\bm{\mathcal{M}}_{\mathrm{obs}}^{(b)}),
        \bm{s}_i^{(b)}
        \right)
            \]
        \ENDFOR
        
        \STATE \textbf{Communication update (defender for $a_i$):}
        \[
        \phi_i \leftarrow \phi_i - \eta_{\phi} \nabla_{\phi_i} 
        \Big(
        \mathcal{L}_{\mathrm{task}}(\bm{\mathcal{M}^{(b)}})
        -
        \beta \mathcal{L}_{\mathrm{rec}}^{(i)}(\bm{\mathcal{M}}_{\mathrm{obs}} ^{(b)}, \bm{s}_i^{(b)})
        \Big) \]
    \ENDFOR
\ENDFOR
\end{algorithmic}
\end{algorithm}

\paragraph{Training details.}
\label{app:training_details}
We use AdamW optimizer with learning rates $\eta_{\phi}$ and $\eta_{\psi}$ for the communication function $g_{ij}$ and decoder $D_i$, respectively. In practice, we perform $k$ steps of decoder updates per communication update to ensure a sufficiently strong adversary. All models are trained on the same dataset splits used for evaluation, and early stopping is applied based on validation performance on task accuracy and ASR. All experiments are run on NVIDIA H100 GPUs. Depending on the backbone model size and communication topology, training typically requires approximately $3$-$8$ hours per configuration under the full-system LCGuard setting.

\subsection{Ablation Studies}
\label{app:ablation}
\subsubsection{Effect of the Tradeoff Parameter $\beta$}
\label{app:effect-of-beta}

The parameter $\beta$ in Eq.~\eqref{eq:minimax} controls the relative strength of leakage suppression against task utility. Smaller values of $\beta$ place more weight on preserving task-relevant information in $\bm{\mathcal{M}}$, while larger values more strongly penalize reconstructable sensitive information. Table~\ref{tab:beta_tradeoff} reports the privacy-utility tradeoff for Full-System LCGuard under different values of $\beta$.

\begin{table}[t]
\centering
\caption{
Effect of the tradeoff parameter $\beta$ for Full-System LCGuard. Increasing $\beta$ improves Privacy Score and lowers Leak Rate and ASR, but excessive values reduce Task Accuracy and Helpfulness by suppressing task-relevant information in the communicated KV artifacts.
}
\small
\setlength{\tabcolsep}{4pt}
\begin{tabular}{c|ccccc}
\toprule
$\bm{\beta}$ 
& \textbf{Privacy} $\uparrow$
& \textbf{Task} $\uparrow$
& \textbf{Help.} $\uparrow$
& \textbf{Leak} $\downarrow$
& \textbf{ASR} $\downarrow$ \\
\midrule
0.00 & 0.455 & \textbf{0.750} & \textbf{0.800} & 0.545 & 0.810 \\
0.05 & 0.690 & 0.745 & 0.775 & 0.310 & 0.520 \\
0.10 & 0.760 & 0.738 & 0.755 & 0.240 & 0.410 \\
0.25 & 0.815 & 0.730 & 0.730 & 0.185 & 0.320 \\
0.50 & 0.845 & 0.715 & 0.700 & 0.155 & 0.270 \\
1.00 & 0.870 & 0.685 & 0.645 & 0.130 & 0.230 \\
2.00 & \textbf{0.895} & 0.610 & 0.535 & \textbf{0.105} & \textbf{0.195} \\
\bottomrule
\end{tabular}
\label{tab:beta_tradeoff}
\end{table}

As $\beta$ increases, LCGuard places greater emphasis on making $\bm{s}_i$ difficult to reconstruct from $\bm{\mathcal{M}}_{\mathrm{obs}}$. This produces a monotonic reduction in Leak Rate and ASR. However, the same transformation can also remove information useful for downstream prediction, so Task Accuracy and Helpfulness gradually decrease. The intermediate range, particularly $\beta \in [0.25,0.50]$, provides the most favorable operating region: ASR is substantially reduced relative to unprotected KV sharing, while helpfulness remains close to the high-utility regime. Very large values of $\beta$ over-prioritize privacy and begin to resemble noise-based defenses, where leakage is reduced at the cost of degraded task performance. 

Figure~\ref{fig:beta_tradeoff} further illustrates this tradeoff, showing a smooth transition along the privacy-utility frontier as $\beta$ increases.

\begin{figure}[t]
    \centering
    \includegraphics[width=0.9\linewidth]{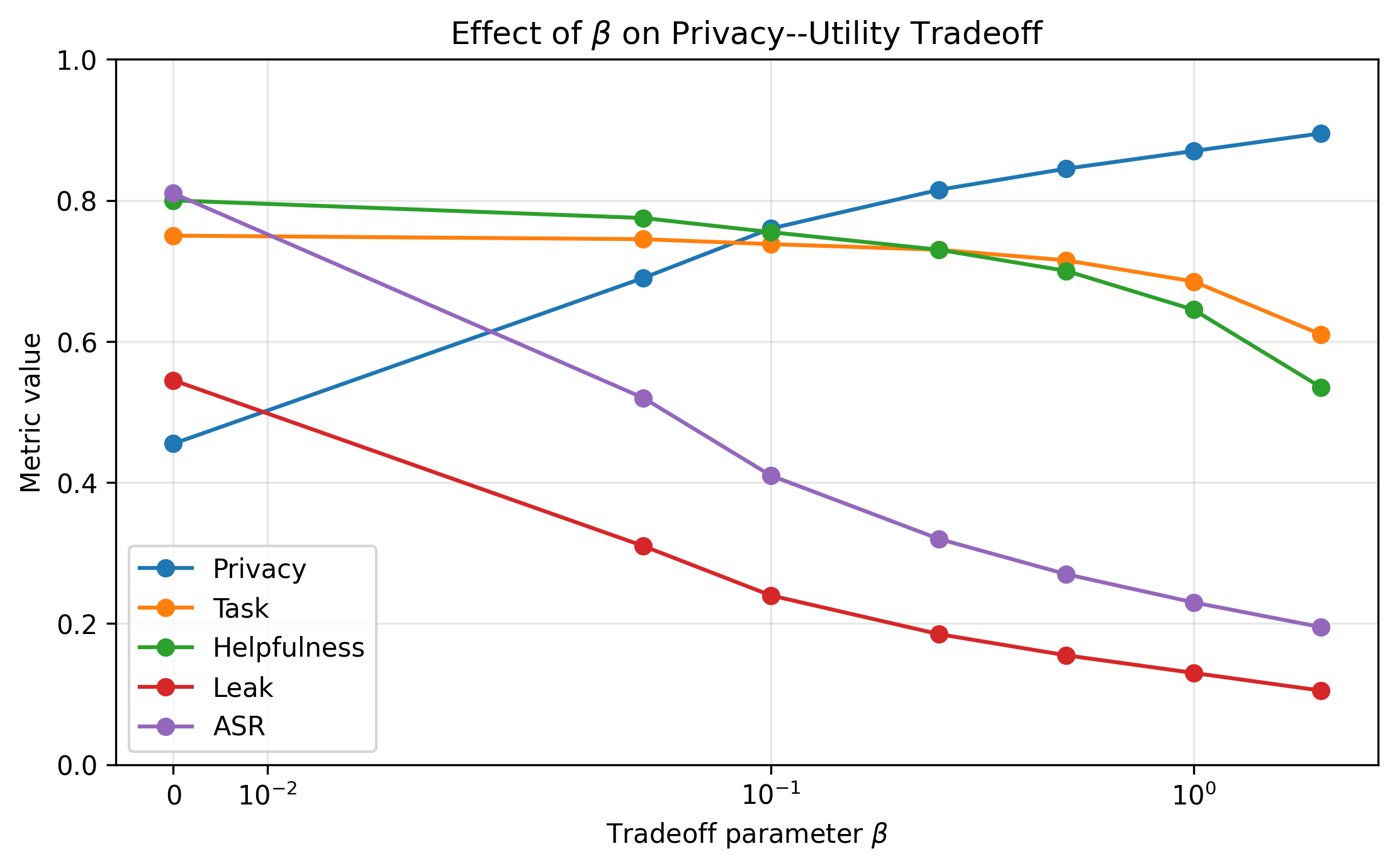}
    \caption{
    Effect of $\beta$ on the privacy-utility tradeoff for \textbf{Full-System LCGuard} on \textbf{Qwen3-4B} under \textbf{sequential communication (4 agents)} on the \textbf{PrivacyLens benchmark}. Increasing $\beta$ monotonically improves Privacy while reducing Leak and ASR, but leads to gradual degradation in Task Accuracy and Helpfulness. The intermediate range $\beta \in [0.25, 0.50]$ provides the best tradeoff between strong privacy and high utility.
    }
    \label{fig:beta_tradeoff}
\end{figure}

\subsubsection{Effect of Adversary Strength}
\label{app:adversary_strength}

To evaluate robustness under stronger threat models, we vary the capacity of the adversarial decoder used to reconstruct sensitive inputs from $\bm{\mathcal{M}}_{\mathrm{obs}}$.

\paragraph{Adversary configurations.}
We consider three adversary architectures of increasing strength:
\begin{itemize}[leftmargin=*]
    \item \textbf{Weak adversary:} shallow MLP with limited capacity. 
    A 2-layer feedforward network with hidden dimension $256$, operating on mean-pooled representations of $\bm{\mathcal{M}}_{\mathrm{obs}}$.

    \item \textbf{Moderate adversary:} multi-layer Transformer decoder. 
    A 2-layer Transformer with hidden dimension $512$ and $8$ attention heads.

    \item \textbf{Strong adversary:} deeper Transformer with larger hidden dimension and attention layers. 
    A 4-layer Transformer with hidden dimension $768$, $12$ attention heads, and feedforward size $2048$.
\end{itemize}

Table~\ref{tab:adversary_strength} reports results for \textbf{Qwen3-4B under sequential communication (4 agents) on PrivacyLens} using Full-System LCGuard with $\beta = 0.25$.

\begin{table}[t]
\centering
\caption{
Effect of adversary strength on \textbf{Full-System LCGuard} for \textbf{Qwen3-4B} under \textbf{sequential communication (4 agents)} on \textbf{PrivacyLens} with $\beta=0.25$. Stronger adversaries achieve higher reconstruction success (higher ASR), but LCGuard maintains significantly lower leakage compared to unprotected KV sharing.
}
\small
\setlength{\tabcolsep}{4pt}
\begin{tabular}{l|ccccc}
\toprule
\textbf{Adversary} 
& \textbf{Privacy} $\uparrow$
& \textbf{Task} $\uparrow$
& \textbf{Help.} $\uparrow$
& \textbf{Leak} $\downarrow$
& \textbf{ASR} $\downarrow$ \\
\midrule
Weak        & 0.835 & 0.730 & 0.730 & 0.165 & 0.285 \\
Moderate & 0.815 & 0.730 & 0.730 & 0.185 & 0.320 \\
Strong  & 0.790 & 0.730 & 0.725 & 0.210 & 0.365 \\
\bottomrule
\end{tabular}
\label{tab:adversary_strength}
\end{table}

As the adversary becomes stronger, reconstruction becomes easier, leading to higher ASR and Leak Rate. However, LCGuard maintains substantial protection even against strong adversaries, indicating that the learned transformation reduces the intrinsic recoverability of sensitive information. Importantly, task utility remains stable across adversary strengths, confirming that robustness improvements primarily affect the privacy dimension without degrading performance.
\subsubsection{Reconstruction Difficulty Analysis}
\label{app:reconstruction_difficulty_ablation}

We further analyze \emph{Reconstruction Difficulty (RD)} as a diagnostic measure of how difficult it is for an adversary to recover sensitive inputs from communicated representations captured through $\mathcal{L}_{\mathrm{prior}}^{(i)}
-
\mathcal{L}_{\mathrm{rec}}^{(i)}(\bm{\mathcal{M}}_{\mathrm{obs}})$. Table~\ref{tab:rd_ablation} summarizes RD alongside ASR and helpfulness across representative models and settings.
\begin{table}[t] \centering
\caption{ Reconstruction difficulty (RD), helpfulness, and ASR across representative models, communication topologies, and agent configurations. Higher RD indicates that reconstruction from communicated KV artifacts approaches prior-level uncertainty. Full-System LCGuard achieves consistently high RD with low ASR while preserving strong helpfulness, demonstrating effective suppression of recoverable sensitive information.}
\small \setlength{\tabcolsep}{3pt} \begin{tabular}{llcc|l|ccc} \toprule \textbf{Model} & \textbf{Topology} & \textbf{\#Agents} & \textbf{Benchmark} & \textbf{Method} & \textbf{RD} $\uparrow$ & \textbf{Help.} $\uparrow$ & \textbf{ASR} $\downarrow$ \\ \midrule Qwen3-4B & Seq & 4 & PrivacyLens & Vanilla KV & 0.31 & \textbf{0.780} & 0.871 \\ & & & & PrivAct & 0.36 & 0.612 & 0.845 \\ & & & & ADAPT & \underline{0.79} & 0.285 & 0.332 \\ & & & & Per-Agent & 0.74 & 0.650 & \underline{0.259} \\ & & & & Full-System & \textbf{0.82} & \underline{0.710} & \textbf{0.216} \\ \midrule Qwen3-8B & Hier & 5 & PrivacyLens & Vanilla KV & 0.32 & \textbf{0.775} & 0.835 \\ & & & & PrivAct & 0.39 & 0.400 & 0.790 \\ & & & & ADAPT & \underline{0.74} & 0.285 & 0.455 \\ & & & & Per-Agent & 0.70 & 0.615 & \underline{0.390} \\ & & & & Full-System & \textbf{0.78} & \underline{0.725} & \textbf{0.345} \\ \midrule Gemma-9B & Seq & 3 & AgentLeak & Vanilla KV & 0.27 & \textbf{0.785} & 0.895 \\ & & & & PrivAct & 0.38 & 0.455 & 0.770 \\ & & & & ADAPT & \underline{0.77} & 0.285 & 0.400 \\ & & & & Per-Agent & 0.73 & 0.660 & \underline{0.275} \\ & & & & Full-System & \textbf{0.83} & \underline{0.720} & \textbf{0.235} \\ \bottomrule \end{tabular}

\label{tab:rd_ablation} \end{table}

\paragraph{Reconstruction difficulty vs. utility.}
Figure~\ref{fig:rd_helpfulness_tradeoff} illustrates the tradeoff between RD and helpfulness. Vanilla KV lies in the high-helpfulness but low-RD region, indicating strong utility but high leakage. ADAPT achieves high RD but at the cost of significant utility degradation. LCGuard shifts the Pareto frontier toward higher RD while preserving substantially higher helpfulness.

\paragraph{Effect of topology and model capacity.}
Reconstruction difficulty under Vanilla KV decreases as communication complexity increases, indicating that compositional interactions amplify leakage. LCGuard mitigates this effect, maintaining high RD across settings. Additionally, larger models tend to exhibit lower RD under Vanilla KV, suggesting that higher-capacity representations encode more recoverable information.

\begin{figure}[t]
    \centering
    \includegraphics[width=0.8\linewidth]{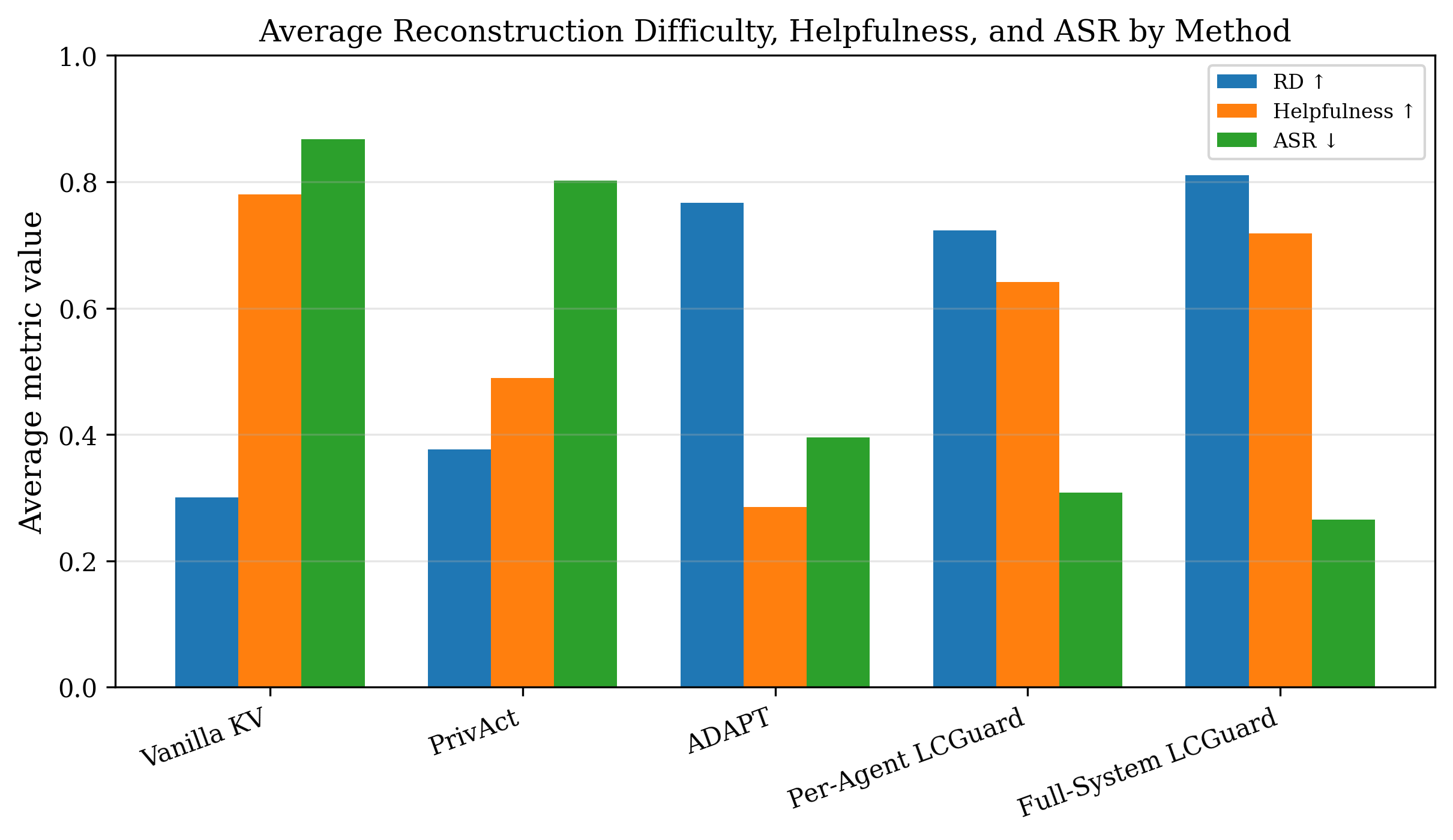}
    \caption{
    Average reconstruction difficulty, helpfulness, and ASR across methods. LCGuard achieves high RD and low ASR while preserving strong helpfulness, unlike ADAPT which sacrifices utility.
    }
    \label{fig:rd_helpfulness_tradeoff}
\end{figure}

\subsection{Additional Results and Analysis}
\label{app:additional_results}

Tables~\ref{tab:qwen8b_additional}, \ref{tab:qwen14b_seq_additional}, and \ref{tab:llama3b_additional} extend the main results to additional model scales, benchmarks, and communication topologies. These results reinforce the core findings while revealing several additional insights regarding model capacity, topology complexity, and communication structure.

\paragraph{Scaling behavior across model sizes.}
Comparing Qwen3-8B (Table~\ref{tab:qwen8b_additional}) and Qwen3-14B (Table~\ref{tab:qwen14b_seq_additional}) with smaller model Qwen3-4B in the Section \ref{sec:results} shows a consistent scaling trend: larger models improve task accuracy and helpfulness under Vanilla KV sharing but also amplify reconstruction risk. For example, in sequential MAGPIE (Table~\ref{tab:qwen14b_seq_additional}), helpfulness reaches $0.840$ while ASR increases to $0.855$, indicating that higher-capacity KV representations encode richer but more recoverable information.

Despite this increased risk, LCGuard remains effective across scales. Full-System LCGuard consistently reduces ASR by a large margin (e.g., $0.855 \rightarrow 0.260$ in Table~\ref{tab:qwen14b_seq_additional}) while preserving most of the utility. This suggests that LCGuard scales favorably with model size and continues to suppress sensitive information even as representation capacity grows.

\paragraph{Effect of communication topology.}
Table~\ref{tab:qwen8b_additional} highlights the impact of topology by comparing sequential, hierarchical, and general graph settings. Moving from sequential to graph-based communication increases ASR for all methods under Vanilla KV (e.g., $0.810 \rightarrow 0.870$ on PrivacyLens), reflecting the greater number of propagation paths and opportunities for compositional leakage.

LCGuard maintains robustness across these increasingly complex settings. While ASR increases slightly in graph topologies, Full-System LCGuard consistently remains the lowest among all methods (e.g., $0.320$ (Seq) vs. $0.370$ (Graph)), indicating that system-level optimization effectively mitigates multi-path leakage. In contrast, Per-Agent LCGuard shows a larger degradation, confirming that local sanitization is insufficient when information propagates through multiple interacting paths.

\paragraph{Dataset-dependent behavior.}
Across all tables, MAGPIE \cite{juneja2025magpie} exhibits higher helpfulness but also stronger coupling between task performance and leakage compared to AgentLeak. For instance, in Table~\ref{tab:llama3b_additional}, Vanilla KV achieves helpfulness $0.760$ on MAGPIE versus $0.750$ on AgentLeak, while maintaining similarly high ASR. This reflects the fact that in MAGPIE, private information is directly tied to task success, making it harder to remove without affecting utility.

LCGuard adapts to this setting by preserving task-relevant signals while reducing reconstruction pathways. Although privacy gains are slightly smaller compared to AgentLeak, Full-System LCGuard still achieves the best tradeoff (e.g., ASR $0.780 \rightarrow 0.290$ in Table~\ref{tab:llama3b_additional}), demonstrating its ability to handle tightly coupled privacy-task dependencies.

\paragraph{Behavior under weaker models.}
Results for LLaMA-3B (Table~\ref{tab:llama3b_additional}) show that smaller models exhibit lower helpfulness and slightly lower ASR under Vanilla KV compared to larger backbones. However, the relative behavior of all methods remains unchanged. LCGuard continues to provide consistent improvements, reducing ASR while maintaining competitive utility.

A notable observation is that the gap between Per-Agent and Full-System LCGuard is smaller for weaker models, suggesting that compositional leakage becomes more pronounced as model capacity increases. This further supports the need for system-level protection in larger, more expressive multi-agent systems.

Overall, the additional results strengthen three key conclusions: 
(1) increasing model capacity amplifies both utility and leakage; 
(2) communication topology directly affects leakage through multi-path propagation; and 
(3) LCGuard remains robust across scales, datasets, and topologies, with Full-System LCGuard consistently achieving the best privacy-utility tradeoff.
\begin{table}[t]
\centering
\caption{
Additional Qwen3-8B results across sequential, hierarchical, and general graph communication topologies. General graph settings use five agents. Leak rate is computed as $1-\mathrm{Privacy}$. Full-System LCGuard consistently achieves the lowest ASR while preserving substantially higher helpfulness than ADAPT.
}
\small
\setlength{\tabcolsep}{2pt}
\begin{tabular}{llc|l|ccccc}
\toprule
\textbf{Topology} & \textbf{Benchmark} & \textbf{\#Agents} & \textbf{Method} 
& \textbf{Privacy} $\uparrow$ 
& \textbf{Task} $\uparrow$
& \textbf{Help.} $\uparrow$
& \textbf{Leak} $\downarrow$
& \textbf{ASR} $\downarrow$ \\
\midrule

\multirow{10}{*}{Seq}

& \multirow{5}{*}{PrivacyLens}
& \multirow{5}{*}{4}
& Vanilla KV Sharing (LatentMAS)  & 0.455 & \textbf{0.750} & \textbf{0.800} & 0.545 & 0.810 \\
& & 
& PrivAct                         & 0.815 & 0.710 & 0.420 & 0.185 & 0.770 \\
& & 
& ADAPT                           & \textbf{0.865} & 0.445 & 0.300 & \textbf{0.135} & 0.430 \\
& & 
& Per-Agent LCGuard               & 0.795 & 0.700 & 0.650 & 0.205 & \underline{0.365} \\
& & 
& Full-System LCGuard             & \underline{0.840} & \underline{0.730} & \underline{0.730} & \underline{0.160} & \textbf{0.320} \\

\cmidrule(lr){2-9}

& \multirow{5}{*}{AgentLeak}
& \multirow{5}{*}{3}
& Vanilla KV Sharing (LatentMAS)  & 0.285 & \textbf{0.670} & \textbf{0.800} & 0.715 & 0.810 \\
& & 
& PrivAct                         & 0.590 & 0.565 & 0.420 & 0.410 & 0.770 \\
& & 
& ADAPT                           & \underline{0.835} & 0.355 & 0.300 & \underline{0.165} & 0.430 \\
& & 
& Per-Agent LCGuard               & 0.790 & 0.615 & 0.650 & 0.210 & \underline{0.365} \\
& & 
& Full-System LCGuard             & \textbf{0.850} & \underline{0.655} & \underline{0.730} & \textbf{0.150} & \textbf{0.320} \\

\midrule

\multirow{10}{*}{Hier}

& \multirow{5}{*}{PrivacyLens}
& \multirow{5}{*}{5}
& Vanilla KV Sharing (LatentMAS)  & 0.430 & \textbf{0.740} & \textbf{0.775} & 0.570 & 0.835 \\
& & 
& PrivAct                         & 0.805 & 0.705 & 0.400 & 0.195 & 0.790 \\
& & 
& ADAPT                           & \underline{0.855} & 0.435 & 0.285 & \underline{0.145} & 0.455 \\
& & 
& Per-Agent LCGuard               & 0.800 & 0.695 & 0.615 & 0.200 & \underline{0.390} \\
& & 
& Full-System LCGuard             & \textbf{0.865} & \underline{0.720} & \underline{0.725} & \textbf{0.135} & \textbf{0.345} \\

\cmidrule(lr){2-9}

& \multirow{5}{*}{AgentLeak}
& \multirow{5}{*}{5}
& Vanilla KV Sharing (LatentMAS)  & 0.265 & \textbf{0.660} & \textbf{0.775} & 0.735 & 0.835 \\
& & 
& PrivAct                         & 0.575 & 0.585 & 0.400 & 0.425 & 0.790 \\
& & 
& ADAPT                           & \underline{0.825} & 0.340 & 0.285 & \underline{0.175} & 0.455 \\
& & 
& Per-Agent LCGuard               & 0.792 & 0.605 & 0.615 & 0.208 & \underline{0.390} \\
& & 
& Full-System LCGuard             & \textbf{0.855} & \underline{0.635} & \underline{0.725} & \textbf{0.145} & \textbf{0.345} \\

\midrule

\multirow{10}{*}{Graph}

& \multirow{5}{*}{PrivacyLens}
& \multirow{5}{*}{5}
& Vanilla KV Sharing (LatentMAS)  & 0.405 & \textbf{0.725} & \textbf{0.745} & 0.595 & 0.870 \\
& & 
& PrivAct                         & 0.785 & 0.690 & 0.375 & 0.215 & 0.835 \\
& & 
& ADAPT                           & \underline{0.835} & 0.415 & 0.255 & \underline{0.165} & 0.490 \\
& & 
& Per-Agent LCGuard               & 0.775 & 0.680 & 0.655 & 0.225 & \underline{0.420} \\
& & 
& Full-System LCGuard             & \textbf{0.845} & \underline{0.705} & \underline{0.705} & \textbf{0.155} & \textbf{0.370} \\

\cmidrule(lr){2-9}

& \multirow{5}{*}{AgentLeak}
& \multirow{5}{*}{5}
& Vanilla KV Sharing (LatentMAS)  & 0.240 & \textbf{0.645} & \textbf{0.745} & 0.760 & 0.870 \\
& & 
& PrivAct                         & 0.550 & 0.565 & 0.375 & 0.450 & 0.835 \\
& & 
& ADAPT                           & \underline{0.805} & 0.325 & 0.255 & \underline{0.195} & 0.490 \\
& & 
& Per-Agent LCGuard               & 0.770 & 0.590 & 0.655 & 0.230 & \underline{0.420} \\
& & 
& Full-System LCGuard             & \textbf{0.830} & \underline{0.625} & \underline{0.705} & \textbf{0.170} & \textbf{0.370} \\

\bottomrule
\end{tabular}
\label{tab:qwen8b_additional}
\end{table}

\begin{table}[t]
\centering
\caption{
Qwen3-14B results on MAGPIE and AgentLeak under sequential communication using KVComm for latent communication. Larger model capacity improves helpfulness but also increases reconstruction risk under Vanilla KV. LCGuard significantly reduces ASR while maintaining strong task performance, with Full-System LCGuard achieving the best privacy-utility tradeoff.
}
\small
\setlength{\tabcolsep}{2pt}
\begin{tabular}{llc|l|ccccc}
\toprule
\textbf{Topology} & \textbf{Benchmark} & \textbf{\#Agents} & \textbf{Method} 
& \textbf{Privacy} $\uparrow$ 
& \textbf{Task} $\uparrow$
& \textbf{Help.} $\uparrow$
& \textbf{Leak} $\downarrow$
& \textbf{ASR} $\downarrow$ \\
\midrule

\multirow{10}{*}{Seq}

& \multirow{5}{*}{MAGPIE}
& \multirow{5}{*}{4}
& Vanilla KV Sharing (KVComm)  & 0.460 & \textbf{0.780} & \textbf{0.840} & 0.540 & 0.855 \\
& & 
& PrivAct                      & 0.830 & 0.740 & 0.655 & 0.170 & 0.800 \\
& & 
& ADAPT                        & \textbf{0.880} & 0.470 & 0.320 & \textbf{0.120} & 0.360 \\
& & 
& Per-Agent LCGuard            & 0.810 & 0.725 & 0.710 & 0.190 & \underline{0.300} \\
& & 
& Full-System LCGuard          & \underline{0.855} & \underline{0.755} & \underline{0.780} & \underline{0.145} & \textbf{0.260} \\

\cmidrule(lr){2-9}

& \multirow{5}{*}{AgentLeak}
& \multirow{5}{*}{3}
& Vanilla KV Sharing (KVComm)  & 0.290 & \textbf{0.690} & \textbf{0.820} & 0.710 & 0.865 \\
& & 
& PrivAct                      & 0.600 & 0.585 & 0.470 & 0.400 & 0.790 \\
& & 
& ADAPT                        & \underline{0.845} & 0.360 & 0.305 & \underline{0.155} & 0.390 \\
& & 
& Per-Agent LCGuard            & 0.800 & 0.630 & 0.690 & 0.200 & \underline{0.310} \\
& & 
& Full-System LCGuard          & \textbf{0.865} & \underline{0.665} & \underline{0.760} & \textbf{0.135} & \textbf{0.270} \\

\bottomrule
\end{tabular}
\label{tab:qwen14b_seq_additional}
\end{table}

\begin{table}[t]
\centering
\caption{
LLaMA-3B results across sequential, hierarchical, and general graph communication topologies on MAGPIE and AgentLeak using KVComm for latent communication. Compared to larger models, overall helpfulness is lower, but the same trend persists: Vanilla KV maximizes utility with high leakage, ADAPT reduces leakage at the cost of utility, and LCGuard achieves the best privacy-utility tradeoff, with Full-System LCGuard consistently yielding the lowest ASR.
}
\small
\setlength{\tabcolsep}{2pt}
\begin{tabular}{llc|l|ccccc}
\toprule
\textbf{Topology} & \textbf{Benchmark} & \textbf{\#Agents} & \textbf{Method} 
& \textbf{Privacy} $\uparrow$ 
& \textbf{Task} $\uparrow$
& \textbf{Help.} $\uparrow$
& \textbf{Leak} $\downarrow$
& \textbf{ASR} $\downarrow$ \\
\midrule

\multirow{10}{*}{Seq}

& \multirow{5}{*}{MAGPIE}
& \multirow{5}{*}{4}
& Vanilla KV Sharing (KVComm)  & 0.410 & \textbf{0.710} & \textbf{0.760} & 0.590 & 0.780 \\
& & 
& PrivAct                      & 0.780 & 0.680 & 0.540 & 0.220 & 0.720 \\
& & 
& ADAPT                        & \textbf{0.830} & 0.400 & 0.280 & \textbf{0.170} & 0.380 \\
& & 
& Per-Agent LCGuard            & 0.760 & 0.660 & 0.640 & 0.240 & \underline{0.330} \\
& & 
& Full-System LCGuard          & \underline{0.800} & \underline{0.690} & \underline{0.710} & \underline{0.200} & \textbf{0.290} \\

\cmidrule(lr){2-9}

& \multirow{5}{*}{AgentLeak}
& \multirow{5}{*}{3}
& Vanilla KV Sharing (KVComm)  & 0.240 & \textbf{0.620} & \textbf{0.750} & 0.760 & 0.800 \\
& & 
& PrivAct                      & 0.550 & 0.540 & 0.420 & 0.450 & 0.730 \\
& & 
& ADAPT                        & \underline{0.800} & 0.320 & 0.260 & \underline{0.200} & 0.410 \\
& & 
& Per-Agent LCGuard            & 0.750 & 0.580 & 0.630 & 0.250 & \underline{0.340} \\
& & 
& Full-System LCGuard          & \textbf{0.810} & \underline{0.600} & \underline{0.690} & \textbf{0.190} & \textbf{0.300} \\

\midrule

\multirow{10}{*}{Hier}

& \multirow{5}{*}{MAGPIE}
& \multirow{5}{*}{5}
& Vanilla KV Sharing (KVComm)  & 0.390 & \textbf{0.700} & \textbf{0.735} & 0.610 & 0.805 \\
& & 
& PrivAct                      & 0.765 & 0.670 & 0.500 & 0.235 & 0.740 \\
& & 
& ADAPT                        & \underline{0.815} & 0.390 & 0.270 & \underline{0.185} & 0.420 \\
& & 
& Per-Agent LCGuard            & 0.755 & 0.650 & 0.615 & 0.245 & \underline{0.360} \\
& & 
& Full-System LCGuard          & \textbf{0.820} & \underline{0.680} & \underline{0.700} & \textbf{0.180} & \textbf{0.320} \\

\cmidrule(lr){2-9}

& \multirow{5}{*}{AgentLeak}
& \multirow{5}{*}{5}
& Vanilla KV Sharing (KVComm)  & 0.220 & \textbf{0.610} & \textbf{0.730} & 0.780 & 0.820 \\
& & 
& PrivAct                      & 0.530 & 0.555 & 0.395 & 0.470 & 0.750 \\
& & 
& ADAPT                        & \underline{0.785} & 0.310 & 0.255 & \underline{0.215} & 0.430 \\
& & 
& Per-Agent LCGuard            & 0.745 & 0.570 & 0.615 & 0.255 & \underline{0.370} \\
& & 
& Full-System LCGuard          & \textbf{0.805} & \underline{0.590} & \underline{0.690} & \textbf{0.195} & \textbf{0.330} \\

\midrule

\multirow{10}{*}{Graph}

& \multirow{5}{*}{MAGPIE}
& \multirow{5}{*}{5}
& Vanilla KV Sharing (KVComm)  & 0.370 & \textbf{0.690} & \textbf{0.710} & 0.630 & 0.840 \\
& & 
& PrivAct                      & 0.740 & 0.660 & 0.470 & 0.260 & 0.780 \\
& & 
& ADAPT                        & \underline{0.800} & 0.380 & 0.250 & \underline{0.200} & 0.470 \\
& & 
& Per-Agent LCGuard            & 0.730 & 0.640 & 0.630 & 0.270 & \underline{0.400} \\
& & 
& Full-System LCGuard          & \textbf{0.795} & \underline{0.670} & \underline{0.695} & \textbf{0.205} & \textbf{0.350} \\

\cmidrule(lr){2-9}

& \multirow{5}{*}{AgentLeak}
& \multirow{5}{*}{5}
& Vanilla KV Sharing (KVComm)  & 0.210 & \textbf{0.600} & \textbf{0.710} & 0.790 & 0.840 \\
& & 
& PrivAct                      & 0.520 & 0.545 & 0.370 & 0.480 & 0.780 \\
& & 
& ADAPT                        & \underline{0.770} & 0.300 & 0.250 & \underline{0.230} & 0.470 \\
& & 
& Per-Agent LCGuard            & 0.720 & 0.560 & 0.630 & 0.280 & \underline{0.400} \\
& & 
& Full-System LCGuard          & \textbf{0.780} & \underline{0.580} & \underline{0.695} & \textbf{0.220} & \textbf{0.350} \\

\bottomrule
\end{tabular}
\label{tab:llama3b_additional}
\end{table}

\subsection{Inference-Time Efficiency Analysis}
\label{app:efficiency_analysis}

We compare the inference-time efficiency of LCGuard against both conventional text-based multi-agent communication and latent KV sharing using LatentMAS~\cite{latentmas}. In text-based systems, each agent must repeatedly decode natural language outputs and re-encode them for downstream agents, introducing substantial autoregressive decoding overhead. LatentMAS reduces this cost by directly transferring KV cache representations between agents, eliminating repeated text generation during intermediate communication stages.

LCGuard preserves the same latent communication paradigm while introducing lightweight sanitization transformations on communicated KV representations before transmission. Since the transformation operates directly on existing KV tensors and avoids additional autoregressive decoding, the resulting inference overhead remains modest relative to text-mediated communication.

Table~\ref{tab:latency_comparison} summarizes the relative inference efficiency on the AgentLeak benchmark under the full-system setting.

\begin{table}[t]
\centering
\small
\setlength{\tabcolsep}{5pt}
\caption{
Inference-time efficiency comparison on the AgentLeak benchmark for Qwen3-4B for sequential 4 agent setting.
Speedup values are reported relative to text-based multi-agent communication.
}
\label{tab:latency_comparison}
\begin{tabular}{lccc}
\toprule
\textbf{Method}
&
\textbf{Communication}
&
\textbf{Relative Latency}
&
\textbf{Speedup}
\\
\midrule

Text-Based MAS
&
Natural language text
&
$1.00\times$
&
$1.0\times$
\\

Vanilla KV Sharing (LatentMAS)
&
Raw KV caches
&
$0.24\times$
&
$\mathbf{4.1\times}$
\\

LCGuard (Full-System)
&
Sanitized KV caches
&
$0.28\times$
&
$3.6\times$
\\

\bottomrule
\end{tabular}
\end{table}

LCGuard retains most of the inference-time efficiency advantages of latent KV communication while introducing only moderate overhead from the lightweight residual bottleneck sanitization applied before KV transmission. The remaining gap relative to Vanilla KV Sharing primarily arises from the additional KV transformation operations performed during communication.

\subsection{Evaluation Metrics}
\label{app:metrics}

We evaluate both downstream utility and representation-level privacy leakage under the threat model defined in Section~\ref{sec:problem}. All metrics are computed over an evaluation dataset $\mathcal{D} = \{(\{\bm{x}_i^{(\ell)}, \bm{s}_i^{(\ell)}\}_{i=1}^N, \bm{y}^{(\ell)})\}_{\ell=1}^n$, where $\ell$ indexes evaluation instances. For each instance, agents communicate via $\bm{\mathcal{M}}^{(\ell)}$ and produce a prediction $\widehat{\bm{y}}^{(\ell)}$. Metrics are averaged over all $n$ instances. Higher values are better for Task Accuracy, Helpfulness, Privacy Score, and Reconstruction Difficulty, while lower values are better for Leak Rate and Attack Success Rate (ASR).

\paragraph{Task Accuracy.}
Task Accuracy measures correctness of the final system output relative to the ground truth. For instance $\ell$, correctness is defined by an indicator function $\mathbbm{1}\{\widehat{\bm{y}}^{(\ell)} = \bm{y}^{(\ell)}\}$ for discrete tasks, or by a normalized task-specific score for structured outputs. We report:
\begin{equation}
\mathrm{Task}
=
\frac{1}{n}
\sum_{\ell=1}^{n}
\mathrm{Score}\big(\widehat{\bm{y}}^{(\ell)}, \bm{y}^{(\ell)}\big),
\end{equation}
where $\mathrm{Score}(\cdot,\cdot) \in [0,1]$ reduces to the indicator function for exact-match tasks.

\paragraph{Helpfulness.}
Helpfulness measures the quality and usefulness of the generated response for completing the task. For each instance $\ell$, a normalized score $h^{(\ell)} \in [0,1]$ is assigned based on either benchmark-provided annotations or evaluation using an LLM-as-a-judge protocol. The overall helpfulness is:
\begin{equation}
\mathrm{Helpfulness}
=
\frac{1}{n}
\sum_{\ell=1}^{n}
h^{(\ell)}.
\end{equation}
While Task Accuracy captures correctness, Helpfulness captures completeness and quality of responses, particularly in open-ended tasks.
\paragraph{Privacy Score.}
Privacy Score measures the extent to which sensitive information $\bm{s}_i$ is not exposed through the system. For each instance $\ell$, let $p^{(\ell)} \in [0,1]$ denote a benchmark-specific privacy score that quantifies whether sensitive information is preserved, computed following the privacy evaluation protocol defined in PrivAct \cite{cheng2026privact}. We report:
\begin{equation}
\mathrm{Privacy}
=
\frac{1}{n}
\sum_{\ell=1}^{n}
p^{(\ell)}.
\end{equation}
Higher values indicate stronger preservation of sensitive information.

\paragraph{Leak Rate.}
Leak Rate quantifies the frequency of sensitive information exposure. Consistent with the Privacy Score definition, we compute:
\begin{equation}
\mathrm{Leak}
=
1 - \mathrm{Privacy}.
\end{equation}
Thus, lower Leak Rate corresponds to better privacy protection.
\paragraph{Attack Success Rate (ASR).}
ASR measures the empirical success of the adversary in recovering sensitive information from communicated artifacts. For each instance $\ell$ and target agent $a_i$, let $\widehat{\bm{s}}_i^{(\ell)}$ be the reconstruction produced by the adversarial decoder $D_i$ given $\bm{\mathcal{M}}_{\mathrm{obs}}^{(\ell)}$, and let $\bm{s}_i^{(\ell)}$ be the true sensitive input. To evaluate semantic equivalence between the reconstructed output and the ground truth, we employ an LLM-as-a-judge framework, following the methodology of previous privacy evaluation benchmarks~\citep{meisenbacher2024llmjudgeprivacy}. Specifically, we prompt a judge LLM to compare the reconstructed and ground-truth sensitive inputs and output a binary decision indicating whether they are semantically equivalent. The attack success indicator for instance $\ell$ is therefore defined as:
\begin{equation}
z^{(\ell)} = \mathbbm{1}\bigl[\text{LLM-Judge}(\widehat{\bm{s}}_i^{(\ell)}, \bm{s}_i^{(\ell)}) = 1\bigr].
\end{equation}
The ASR is then the average over all instances:
\begin{equation}
\mathrm{ASR} = \frac{1}{n} \sum_{\ell=1}^{n} z^{(\ell)}.
\end{equation}.

\paragraph{Reconstruction Difficulty.}
Reconstruction Difficulty is derived from the reconstruction loss $\mathcal{L}_{\mathrm{rec}}^{(i)}$ and prior loss $\mathcal{L}_{\mathrm{prior}}^{(i)}$ defined in Section~\ref{sec:problem}. For each agent $a_i$, we define:
\begin{equation}
\Delta_i
=
\frac{
\mathcal{L}_{\mathrm{prior}}^{(i)}
-
\mathcal{L}_{\mathrm{rec}}^{(i)}(\bm{\mathcal{M}}_{\mathrm{obs}})
}{
\mathcal{L}_{\mathrm{prior}}^{(i)}
}.
\label{equ:rec_difficulty}
\end{equation}
This measures how much additional information about $\bm{s}_i$ is revealed by the communicated artifacts relative to the prior. Values close to $0$ indicate that $\bm{\mathcal{M}}_{\mathrm{obs}}$ provides little information beyond the prior (high reconstruction difficulty, strong privacy), while values near $1$ indicate near‑perfect reconstruction (low difficulty, high leakage). We report the average across agents:
\begin{equation}
\Delta
=
\frac{1}{N}
\sum_{i=1}^{N}
\Delta_i.
\end{equation}

\subsection{Detailed Experimental Setup}
\label{app:exp_details}

We provide additional details on agent configurations, communication structure, architecture choices, and training hyperparameters used in the experiments. All components follow the formulation in Section~\ref{sec:problem} and Section~\ref{sec:method}.

\paragraph{Agent configurations and roles.}
\label{par:agent_rol}
We instantiate multi-agent systems with $N \in \{3,4\}$ agents for sequential and hierarchical topologies, and $N=5$ agents for general graph settings. Each agent $a_i$ receives task input $\bm{x}_i$ and agent-specific sensitive input $\bm{s}_i$, and produces KV representations $(\bm{K}_i,\bm{V}_i)$. All agents share the same backbone architecture but differ in inputs and communication context.

\paragraph{Sequential architectures ($N \in \{3,4\}$).}
Agents are arranged in a directed chain $a_1 \rightarrow a_2 \rightarrow \dots \rightarrow a_N$, where each agent consumes upstream representations and produces $\bm{m}_{ii+1}$.

\begin{itemize}[leftmargin=*]
    \item \textbf{Agent $a_1$ (Input Encoder):} processes the initial task input $\bm{x}_1$ and produces the base representation $(\bm{K}_1,\bm{V}_1)$.
    
    \item \textbf{Agent $a_2$ (Primary Reasoner):} consumes $\bm{m}_{12}$ and performs core task reasoning, generating structured intermediate representations.
    
    \item \textbf{Agent $a_3$ (Context Refiner, when $N=4$):} integrates upstream representations with local inputs $(\bm{x}_3,\bm{s}_3)$ to refine and stabilize the latent state.
    
    \item \textbf{Final agent ($a_3$ for $N=3$, $a_4$ for $N=4$) (Output Generator):} aggregates all upstream information and produces the final representation used to predict $\bm{y}$.
\end{itemize}

This configuration reflects staged reasoning, where information is progressively transformed. The chain structure naturally induces \emph{progressive exposure}, since sensitive information encoded early can propagate and become increasingly recoverable downstream.

\paragraph{Hierarchical architectures ($N \in \{3,4\}$).}
Agents are organized into two levels: multiple leaf agents feeding into a higher-level aggregator.

\begin{itemize}[leftmargin=*]
    \item \textbf{Leaf agents (e.g., $a_1,a_2$ or $a_1,a_2,a_3$):} independently process different portions of the task input and generate latent representations $(\bm{K}_i,\bm{V}_i)$ from $(\bm{x}_i,\bm{s}_i)$.
    
    \item \textbf{Aggregator agent (root):} receives $\{\bm{m}_{ij}\}$ from all leaf agents and integrates them to produce the final system output $\bm{y}$.
\end{itemize}

Unlike sequential pipelines, this structure enables \emph{parallel information flow} followed by aggregation.

\paragraph{General graph architectures ($N=5$).}
We consider a fixed 5-agent communication graph that captures  multi-path information flow. 

The interaction pattern is defined as follows:
\begin{itemize}[leftmargin=*]
    \item \textbf{Agents $a_1, a_2$ (source agents):} primarily encode task inputs and initiate information flow.
    
    \item \textbf{Agents $a_3, a_4$ (intermediate agents):} receive representations from multiple upstream agents and propagate transformed representations to downstream agents.
    
    \item \textbf{Agent $a_5$ (sink/aggregator):} collects representations from multiple paths and produces the final output $\bm{y}$.
\end{itemize}

Such a topology captures realistic collaborative systems where information is processed along multiple routes before aggregation. This significantly increases the difficulty of controlling leakage, since sensitive signals can be:
(i) propagated along different paths,
(ii) recombined at intermediate agents, and
(iii) amplified at the final aggregator.

% \begin{figure}[t]
% \centering
% \begin{tikzpicture}[
%     node distance=2cm,
%     every node/.style={draw, circle, minimum size=0.9cm}
% ]
% \node (a1) {$a_1$};
% \node (a2) [below of=a1] {$a_2$};
% \node (a3) [right of=a1, xshift=1.5cm] {$a_3$};
% \node (a4) [below of=a3] {$a_4$};
% \node (a5) [right of=a3, xshift=2cm] {$a_5$};

% \draw[->] (a1) -- (a3);
% \draw[->] (a2) -- (a3);
% \draw[->] (a3) -- (a4);
% \draw[->] (a2) -- (a4);
% \draw[->] (a4) -- (a5);
% \draw[->] (a3) -- (a5);

% \end{tikzpicture}
% \caption{General graph topology with $5$ agents. Multiple communication paths enable information propagation and aggregation across the network.}
% \label{fig:graph_topology_appendix}
% \end{figure}
\begin{figure}[t]
\centering
\includegraphics[width=0.82\linewidth]{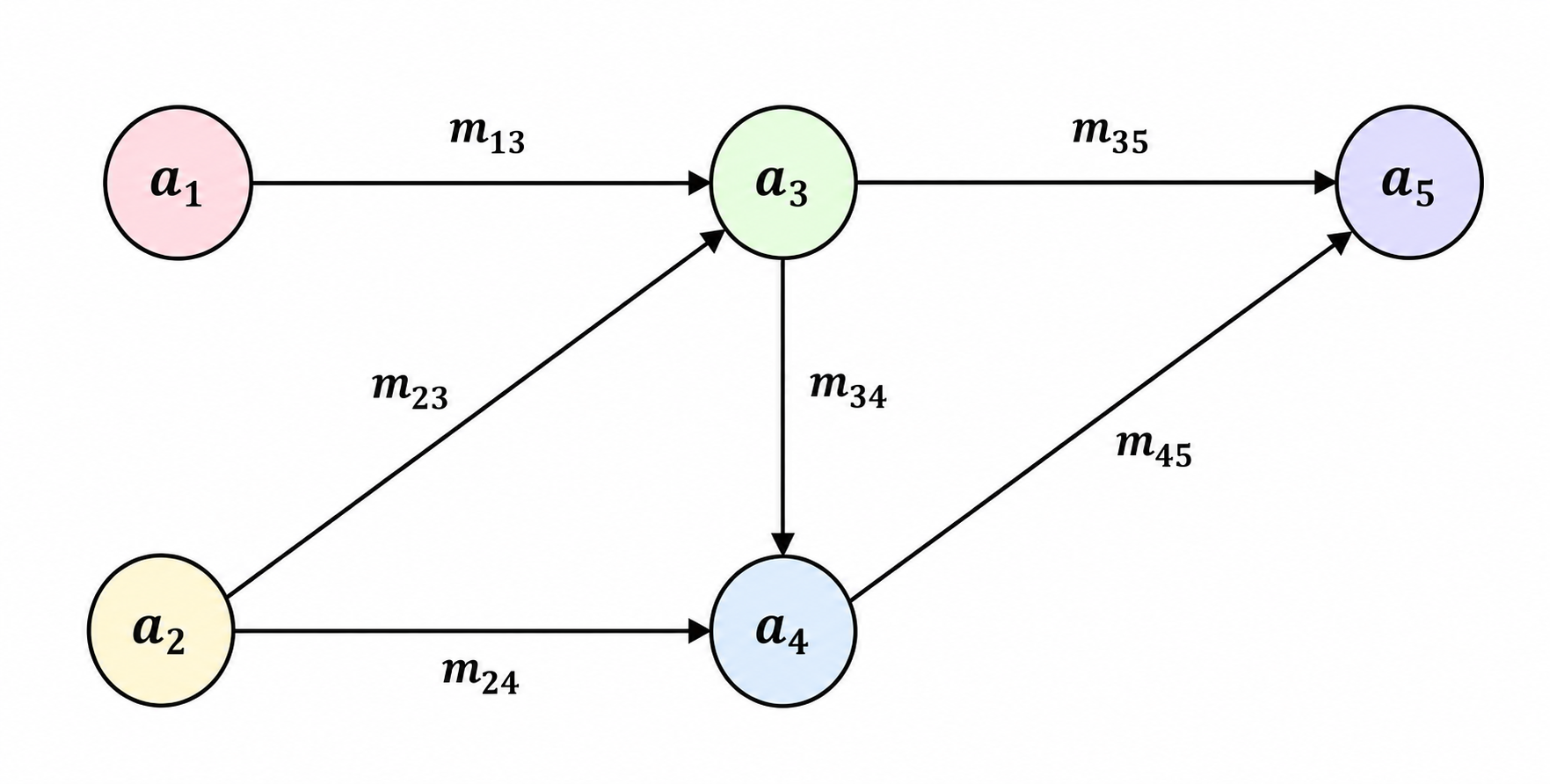}
\caption{
General graph topology with $5$ agents. Multiple communication paths enable information propagation and aggregation across the network.
}
\label{fig:graph_topology_appendix}
\end{figure}
\paragraph{Baselines and comparison setup.}
We compare LCGuard against representative baselines that operate at different levels of the system. \textbf{Vanilla KV Sharing} (LatentMAS \cite{latentmas} or KVComm \cite{shi2026kvcomm}) directly transmits raw KV representations, i.e., $g_{ij}$ is the identity function, providing the best utility but no protection against reconstruction-based leakage. \textbf{PrivAct}~\citep{cheng2026privact} enforces contextual privacy at the policy level by training agents to generate outputs that satisfy privacy preferences. \textbf{ADAPT}~\citep{fatima2025privacymas} applies adaptive differential privacy by injecting Gaussian noise into KV representations before communication, $\bm{\mathcal{C}}^{\mathrm{dp}}_i = \bm{\mathcal{C}}^{\mathrm{raw}}_i + \bm{\xi}_i$ with $\bm{\xi}_i \sim \mathcal{N}(0, \sigma_i^2 \mathbf{I})$, where the noise scale $\sigma_i$ is adjusted via a dynamic privacy budget. This reduces leakage but also perturbs task-relevant signals. Finally, \textbf{Per-Agent LCGuard} applies the proposed transformation $g_{ij}$ independently at each agent, without accounting for system-level aggregation effects. 
\paragraph{Training details.}
We optimize Eq.~\eqref{eq:minimax} using alternating updates between communication parameters $\{\bm{\phi}_i\}$ and adversarial parameters $\{\bm{\psi}_i\}$.

\begin{itemize}[leftmargin=*]
    \item \textbf{Optimizer:} AdamW for both communication functions and decoders.
    \item \textbf{Learning rates:} $\eta_{\phi} = 1 \times 10^{-4}$ for $g_{ij}$, $\eta_{\psi} = 5 \times 10^{-4}$ for $D_i$.
    \item \textbf{Batch size:} $8$ or $16$ depending on model size.
    \item \textbf{Alternation schedule:} one adversary update followed by one communication update per iteration.
\end{itemize}

\paragraph{Tradeoff parameter.}
We treat $\beta$ as a tunable hyperparameter and evaluate multiple values in the range $\beta \in [0.05, 2.0]$. Unless otherwise specified, reported results use $\beta = 0.25$ for Per-Agent LCGuard and $\beta = 0.5$ for Full-System LCGuard, which provide a strong privacy-utility tradeoff (see Appendix~\ref{app:effect-of-beta}).

\paragraph{Evaluation protocol.}
For each benchmark, we construct multi-agent tasks with agent-specific inputs $\{\bm{s}_i\}$. During evaluation:
\begin{itemize}[leftmargin=*]
    \item Agents communicate only via $\bm{\mathcal{M}}$.
    \item The final output $\widehat{\bm{y}}$ is produced from the aggregated representations.
    \item Adversarial decoders are evaluated on held-out data to compute $\mathcal{L}_{\mathrm{rec}}^{(i)}$ and ASR.
\end{itemize}

All reported metrics are averaged over 3 runs. The same communication topology, agent configuration, and data splits are used across all methods to ensure fair comparison.
\paragraph{Inference only.}
During inference, the adversarial decoders are discarded; only the optimised communication functions $\{g_{ij}\}$ are used to transform KV caches before transmission. Hence LCGuard introduces no inference‑time overhead beyond the learned transformations (a single feed‑forward projection per communicated artifact).
\subsubsection{Communication Transformation Architecture}
\label{app:comm_architecture}

LCGuard regulates latent communication through lightweight learnable transformation functions applied directly to shared KV representations before transmission between agents. For each communication edge $(a_i,a_j)$, the communicated artifact is defined as
\begin{equation}
\bm{m}_{ij} = g_{ij}(\bm{K}_i,\bm{V}_i),
\end{equation}
where $\bm{K}_i \in \mathbb{R}^{T_i \times d_k}$ and $\bm{V}_i \in \mathbb{R}^{T_i \times d_v}$ denote the key and value tensors generated by agent $a_i$.

\vspace{2pt}
\noindent
\textbf{Architecture.}
Each communication function $g_{ij}$ is implemented as a lightweight residual bottleneck transformation operating independently on keys and values:
\begin{align}
\bm{K}_i^{\mathrm{san}}
&=
\bm{K}_i
+
W^{K}_{2}
\,\sigma\!\left(
W^{K}_{1}\,
\mathrm{LN}(\bm{K}_i)
\right), \\
\bm{V}_i^{\mathrm{san}}
&=
\bm{V}_i
+
W^{V}_{2}
\,\sigma\!\left(
W^{V}_{1}\,
\mathrm{LN}(\bm{V}_i)
\right),
\end{align}
where $\mathrm{LN}(\cdot)$ denotes layer normalization, $\sigma(\cdot)$ is the GELU activation function, and
\begin{equation}
W^{K}_{1} \in \mathbb{R}^{d_b \times d_k},
\quad
W^{K}_{2} \in \mathbb{R}^{d_k \times d_b},
\end{equation}
with analogous definitions for the value projections. Here $d_b \ll d_k,d_v$ denotes a low-dimensional bottleneck that constrains the amount of information preserved in the communicated representation.

The learnable communication parameters for agent $a_i$ are
\begin{equation}
\bm{\phi}_i
=
\{
W^{K}_{1},W^{K}_{2},
W^{V}_{1},W^{V}_{2}
\}.
\end{equation}

The final communicated artifact transmitted to downstream agents is
\begin{equation}
\bm{m}_{ij}
=
(\bm{K}_i^{\mathrm{san}},\bm{V}_i^{\mathrm{san}}).
\end{equation}

\vspace{2pt}
\noindent
\textbf{Design rationale.}
The residual bottleneck structure serves two complementary purposes. First, the residual connection preserves task-relevant semantic information necessary for downstream reasoning and coordination. Second, the low-dimensional bottleneck restricts the direct propagation of fine-grained agent-specific information, encouraging the transformation to compress communicated representations toward task-relevant features while discarding reconstructable sensitive content.

Unlike noise-based perturbation approaches, the transformation is learned jointly with the adversarial objective in Eq.~\eqref{eq:minimax}, enabling the communicated representations to adaptively balance utility preservation and leakage suppression.

\vspace{2pt}
\noindent
\textbf{Parameterization across agents.}
For the Per-Agent LCGuard setting, each agent maintains an independent communication transformation module parameterized by $\bm{\phi}_i$. In the Full-System LCGuard setting, all communication functions are optimized jointly through the system-level objective in Eq.~\eqref{eq:minimax}, allowing the transformations to account for compositional leakage arising across multiple communication paths and aggregation stages.

\vspace{2pt}
\noindent
\textbf{Implementation details.}
Unless otherwise stated, the bottleneck dimension is set to $d_b = d_k/4$, which provided a favorable balance between utility preservation and leakage suppression in preliminary validation experiments. The transformation is applied to the final-layer KV representations before communication and operates token-wise across the sequence dimension. For hidden dimension $d=4096$ and bottleneck ratio $d_b=d/4$, the communication transformation introduces approximately $16.8$M additional trainable parameters. All transformation parameters are optimized jointly with the adversarial training procedure described in Section~\ref{sec:method}.

\subsection{Baselines}
\label{app:baselines}

We provide detailed descriptions of all baselines used in our evaluation. These methods span complementary approaches to privacy in multi-agent LLM systems, including unprotected latent communication, representation-level perturbation, and policy-level alignment. All baselines operate under the same communication framework defined in Section~\ref{sec:problem}, where agents exchange latent artifacts $\bm{m}_{ij}$ derived from KV representations $(\bm{K}_i,\bm{V}_i)$.

\subsubsection{Vanilla KV Sharing (No Protection)}

This baseline corresponds to standard latent communication (LatentMAS \cite{latentmas} and KVComm \cite{shi2026kvcomm} used in our experiments) without any privacy mechanism. Communication functions are set to identity, i.e.,
\[
\bm{m}_{ij} = g_{ij}(\bm{K}_i,\bm{V}_i) = (\bm{K}_i,\bm{V}_i).
\]
Thus, agents directly share their raw KV representations across communication steps.

While this setting preserves maximal task-relevant information and typically achieves the highest task accuracy and helpfulness, it exposes the full internal representations of each agent. Under the reconstruction-based threat model in Section~\ref{sec:problem}, this leads to high leakage and ASR.

\subsubsection{ADAPT: Adaptive Differential Privacy}

We adopt the ADAPT mechanism from PrivacyMAS~\citep{fatima2025privacymas}, which applies differential privacy through noise injection in multi agent systems. Instead of fixed noise, ADAPT dynamically adjusts the privacy budget based on system behavior, enabling a flexible privacy-utility tradeoff.

In our setting, ADAPT perturbs KV representations before communication:
\begin{equation}
\widetilde{\bm{K}}_i = \bm{K}_i + \bm{\xi}_i^{K}, 
\quad
\widetilde{\bm{V}}_i = \bm{V}_i + \bm{\xi}_i^{V},
\end{equation}
where
\[
\bm{\xi}_i^{K} \sim \mathcal{N}(0, \sigma_i^2 \mathbf{I}), 
\quad
\bm{\xi}_i^{V} \sim \mathcal{N}(0, \sigma_i^2 \mathbf{I}),
\]
and $\sigma_i$ is determined by the adaptive privacy budget. The communicated artifact is then
\[
\bm{m}_{ij} = (\widetilde{\bm{K}}_i,\widetilde{\bm{V}}_i).
\]

This mechanism reduces the amount of recoverable information in $\bm{\mathcal{M}}$ by introducing stochastic perturbations. However, it also degrades task-relevant information, often leading to reduced task accuracy and helpfulness.

\subsubsection{PrivAct: Contextual Privacy Alignment}

PrivAct~\citep{cheng2026privact} is a policy-level approach that incorporates contextual privacy preferences into agent behavior through preference-based learning. Rather than modifying latent communication, PrivAct trains agents to generate outputs that comply with privacy constraints conditioned on context.

In our setting, PrivAct does not modify the communication mechanism, i.e.,
\[
\bm{m}_{ij} = g_{ij}(\bm{K}_i,\bm{V}_i)
\]
remains unchanged (typically identity), and privacy is enforced only at the output level through the learned policy.

As a result, while PrivAct can improve observable privacy metrics (e.g., reducing explicit disclosure in outputs), it does not constrain the information content of $\bm{\mathcal{M}}$. Under our reconstruction-based threat model, latent representations may still encode sensitive information, leading to high ASR despite improved surface-level privacy.

\subsubsection{Per-Agent LCGuard (Local Variant)}

This variant applies LCGuard independently at each agent without system-level coordination. Each agent learns its own communication function $g_{ij}$ (parameterized by $\bm{\phi}_i$) and transforms its KV representations before transmission:
\[
\bm{m}_{ij} = g_{ij}(\bm{K}_i,\bm{V}_i).
\]

The optimization follows the same adversarial formulation as Eq.~\eqref{eq:minimax}, but is applied locally with $\bm{\mathcal{M}}_{\mathrm{obs}}$ restricted to individual communication links (e.g., $\bm{m}_{ij}$). No joint optimization across agents is performed.

This baseline evaluates whether local sanitization alone is sufficient to mitigate leakage. Since it does not account for interactions between multiple communication paths, it cannot fully capture compositional leakage effects that arise at the system level.

\subsection{Related Work}
\label{app:related_work}
\paragraph{Multi-agent LLM communication.}
LLM-based multi-agent systems (MAS) have emerged as a powerful paradigm for extending the reasoning capabilities of individual models through collaboration \cite{li2023camel, wu2024autogen, hong2024metagpt, guo2024llm, tran2025multiagent}. Prior work has explored a wide range of communication strategies, including role-based coordination, debate-style reasoning, and iterative refinement \cite{du2023debate, yang2025agentnet, qian2025scaling}. A central factor underlying the effectiveness of these systems is the communication mechanism between agents. Existing approaches primarily rely on natural language exchange, with recent efforts improving efficiency, topology design, and interaction granularity \cite{du2024learning_multiagent_communication, leong2025amas, zeng2025graph_diffusion}. Despite these advances, communication remains grounded in token-level representations, inheriting the inefficiencies and limitations of text-based interfaces \cite{wang2025agentdropout}.

\paragraph{Latent communication in multi-agent systems.}
To overcome the limitations of text-based interaction, recent work has explored latent communication mechanisms that operate directly on internal model representations. ThoughtComm \cite{zheng2025thought} introduces a shared latent space for communication using encoder–decoder modules, while Cache-to-Cache (C2C) \cite{fu2026cache_to_cache} enables cross-model transfer via KV projections. LatentMAS \cite{latentmas} further proposes propagation of KV-based latent reasoning across agents, demonstrating improved efficiency and reasoning performance. More recent work explores direct KV-based communication and reuse of intermediate attention states across models and agents \cite{shi2026kvcomm, pan2025kvflow, bian2026tokendance}. Other approaches investigate token-level embedding exchange or centralized latent aggregation \cite{du2025latent, liu2026consensus_trap}. These methods establish latent representations, particularly KV caches, as an effective communication substrate. However, they primarily focus on improving expressiveness and efficiency, without explicitly modeling the information content or potential leakage within shared representations.

\paragraph{Privacy and information leakage in LLMs.}
A growing body of work has demonstrated that LLMs and their internal representations can encode and reveal sensitive information. Training-time leakage has been extensively studied through memorization and extraction attacks \cite{carlini2020extracting, nasr2025scalable_extraction, chang2025context_aware_mia, nakka2024pii_compass, yan2024backdooring}, while inference-time attacks show that intermediate representations can be inverted to recover inputs \cite{servedio2025hidden_states, petrov2024dager, gao2024dory}. These findings indicate that model representations retain significant information about their inputs, even when not explicitly exposed in outputs. In multi-agent settings, this raises new concerns, as intermediate representations are reused and propagated across agents, creating additional pathways for information exposure \cite{afzal2025knowing_before_saying, chen2024inside, jiao2026llm_safety_within}.

\paragraph{KV-cache security and reconstruction attacks.}
Recent work has identified KV caches as a critical and underexplored attack surface in LLM inference \cite{luo2026shadow, feng2025identify}. KV caches store intermediate attention states that maintain a strong correspondence with input tokens, making them susceptible to reconstruction attacks. Prior studies demonstrate multiple attack vectors, including inversion-based reconstruction, collision-based matching, and instruction-based extraction \cite{luo2026shadow}. These works show that an adversary with access to KV caches can recover sensitive inputs with high fidelity. Complementary efforts, such as SafeKV \cite{chu2025safekv}, propose system-level defenses through selective cache sharing and isolation in multi-tenant environments. However, these approaches focus on serving-time cache management and do not consider scenarios where KV representations are intentionally transmitted across agents as part of a communication protocol.

Our work lies at the intersection of latent communication and representation-level privacy in MAS. Unlike prior MAS research, we focus on the information encoded in intermediate representations rather than communication protocols or agent design. Unlike latent communication methods, which primarily optimize efficiency and reasoning, we explicitly model and constrain the recoverability of agent-specific inputs from shared representations. Finally, unlike prior KV-cache security work, which treats KV caches as artifacts of inference infrastructure, we consider them as an active communication channel between agents. This perspective introduces a fundamentally different problem: controlling information flow under representation sharing. LCGuard addresses this by learning transformations that preserve task-relevant semantics while limiting reconstruction-based leakage.

\subsection{Limitations}
\label{app:limitations}
LCGuard is evaluated on selected open-weight LLM families, benchmark tasks, and fixed communication topologies, so its behavior may differ under heterogeneous agents, very large-scale deployments, or multimodal latent communication. The framework also assumes access to paired training data for adversarial reconstruction and relies on the strength of the chosen decoder as a proxy for leakage risk. While LCGuard reduces empirical reconstruction success, it does not provide formal privacy guarantees such as differential privacy.

\subsection{Broader Impacts}
\label{app:broader_impacts}
LCGuard has positive potential impact by reducing representation-level privacy leakage in multi-agent LLM systems that share latent KV artifacts. This may support safer deployment of collaborative agents in settings involving sensitive user context, retrieved documents, or private intermediate reasoning. However, reconstruction-based evaluation methods could also inform stronger attacks if misused, and LCGuard should not be interpreted as providing formal privacy guarantees. Responsible deployment should combine representation-level defenses with access control, logging safeguards, output-level privacy checks, and careful evaluation under realistic threat models.

\newpage

\end{document}